\documentclass[10pt,twocolumn,letterpaper]{article}

\usepackage[pagenumbers]{cvpr}      %

\usepackage{bbm}
\usepackage{amsmath}
\usepackage{graphicx}
\usepackage{multirow}
\usepackage{makecell}
\usepackage{blindtext}
\usepackage{pifont}
\usepackage{arydshln}
\usepackage{xhfill}

\newcommand{\cmark}{\ding{51}}%
\newcommand{\xmark}{\ding{55}}

\newcommand{\methodshortname}{RoMo\xspace}
\definecolor{cvprblue}{rgb}{0.21,0.49,0.74}
\definecolor{gold}{rgb}{0.7, 0.5, 0}

\newcommand{\TODO}[1]{\textbf{\color{red}[TODO: #1]}}
\newcommand{\SA}[1]{\textcolor{violet}{[SS: #1]}}

\newcommand{\LILY}[1]{\textcolor{orange}{[LG: #1]}}

\newcommand{\GK}[1]{\textbf{\color{cyan}[GK: #1]}}

\newcommand{\MM}[1]{\textcolor{gold}{[MM: #1]}}

\newcommand{\SRBS}[1]{{\textcolor{purple}{[SRBS: #1]}}}

\newcommand{\DL}[1]{{\color{dark_green}{\bf [DL: #1]}}}

\newcommand{\AT}[1]{{\color{cvprblue}{\bf [AT: #1]}}}

\newcommand{\DF}[1]{{\color{blue}{\bf [DF: #1]}}}

\newcommand{\MAB}[1]{{\color{green}{\bf [MAB: #1]}}}

\newcommand{\AJ}[1]{{\color{red}{\bf [AJ: #1]}}}

\renewcommand{\TODO}[1]{}
\renewcommand{\SA}[1]{}

\renewcommand{\LILY}[1]{}

\renewcommand{\GK}[1]{}

\renewcommand{\MM}[1]{}

\renewcommand{\SRBS}[1]{}

\renewcommand{\DL}[1]{}

\renewcommand{\AT}[1]{}

\renewcommand{\DF}[1]{}

\renewcommand{\MAB}[1]{}

\renewcommand{\AJ}[1]{}

\renewcommand{\paragraph}[1]{\vspace{.5em}\noindent\textbf{#1.}}

\newcommand{\uv}{\mathbf{u}}
\newcommand{\image}{\mathcal{I}}
\newcommand{\mask}{\mathcal{M}}
\newcommand{\dynamics}{\mathcal{D}}
\newcommand{\indicator}{\mathbbm{1}}
\newcommand{\flow}{\mathcal{F}}
\newcommand{\x}{x}

\newcommand{\fundamental}{\mathbf{F}}
\newcommand{\sampson}{\mathbf{S}}
\newcommand{\upper}{\mathbf{U}}
\newcommand{\lowerr}{\mathbf{L}}
\newcommand{\threshup}{\theta_u}
\newcommand{\threshlow}{\theta_l}
\newcommand{\avgflow}{{v_t}}
\newcommand{\steps}{K}
\newcommand{\feat}{\mathbf{G}}
\newcommand{\mlp}{\mathcal{H}}
\newcommand{\param}{\theta}
\newcommand{\loss}{\mathcal{L}}

\makeatletter
\renewcommand{\@makefnmark}{}
\makeatother

\newcount\rowc

\makeatletter
\def\ttabular{%
\hbox\bgroup
\let\\\cr
\def\rulea{\ifnum\rowc=\@ne \hrule height 1.3pt \fi}
\def\ruleb{
\ifnum\rowc=1\hrule height 1.3pt \else
\ifnum\rowc=6\hrule height \heavyrulewidth 
   \else \hrule height \lightrulewidth\fi\fi}
\valign\bgroup
\global\rowc\@ne
\rulea
\hbox to 10em{\strut \hfill##\hfill}%
\ruleb
&&%
\global\advance\rowc\@ne
\hbox to 10em{\strut\hfill##\hfill}%
\ruleb
\cr}
\def\endttabular{%
\crcr\egroup\egroup}

\definecolor{cvprblue}{rgb}{0.21,0.49,0.74}
\usepackage[pagebackref,breaklinks,colorlinks,allcolors=cvprblue]{hyperref}

\def\paperID{16955} %
\def\confName{CVPR}
\def\confYear{2025}

\title{\methodshortname: Robust Motion Segmentation Improves Structure from Motion\\[.1em] {\large\url{https://romosfm.github.io}}} \vspace{-1.1em}

\author{
Lily Goli\textsuperscript{* 1,2} \and
Sara Sabour\textsuperscript{* 1,2} \and
Mark Matthews\textsuperscript{1} \and 
Marcus Brubaker\textsuperscript{1} \and
Dmitry Lagun\textsuperscript{1} \and
Alec Jacobson\textsuperscript{2,3} \and
David J.\ Fleet\textsuperscript{1,2} \and
Saurabh Saxena\textsuperscript{$\dagger$ 1} \and 
Andrea Tagliasacchi\textsuperscript{$\dagger$ 1,2,4}  \and
\ \\
\vspace*{-2em}
\textsuperscript{1}Google DeepMind $\;$
\textsuperscript{2}University of Toronto $\;$
\textsuperscript{3}Adobe Research $\;$
\textsuperscript{4}Simon Fraser University
\vspace{2em}
}

\begin{document}
\maketitle
\begin{abstract}
\vspace*{-0.75cm}

There has been extensive progress in the reconstruction and generation of 4D scenes from monocular casually-captured video. 
While these tasks rely heavily on known camera poses, the problem of finding such poses using structure-from-motion (SfM) often depends on robustly separating static from dynamic parts of a video.
The lack of a robust solution to this problem limits the performance of SfM camera-calibration pipelines.
We propose a novel approach to video-based motion segmentation to identify the components of a scene that are moving w.r.t.\ a fixed world frame.
Our simple but effective iterative method, \methodshortname, combines optical flow and epipolar cues with a pre-trained video segmentation model.
It outperforms unsupervised baselines for motion segmentation as well as supervised baselines trained from synthetic data.
More importantly, the combination of an off-the-shelf SfM pipeline with our segmentation masks establishes a new state-of-the-art on camera calibration for scenes with dynamic content,
outperforming existing methods by a substantial margin.
\end{abstract}
    
\vspace*{-0.15cm}
\section{Introduction}
\label{sec:intro}
\vspace*{-0.1cm}

The segmentation of moving objects in video, i.e. disentangling object motion from camera-induced motion, is a natural precursor to myriad downstream tasks and applications, including augmented reality~\cite{Hammer2016msa}, autonomous navigation~\cite{Klappstein2009mos, Rashed2019mda}, action recognition~\cite{Weinland2011svb} and 4D scene reconstruction~\cite{Wang2021dymslam}.
In this paper, we are particularly interested in motion segmentation as a means of improving the robustness of structure-from-motion (SfM) methods (e.g., COLMAP~\cite{schoenberger2016mvs, schoenberger2016sfm}).
Moving objects are problematic, as they violate the SfM rigidity assumption, greatly limiting the videos to which SfM can be successfully applied.
\footnote{$^{*}$ Equal contribution}
\footnote{$^\dagger$ Equal advising}

\begin{figure}
\centering
\includegraphics[width=\linewidth]{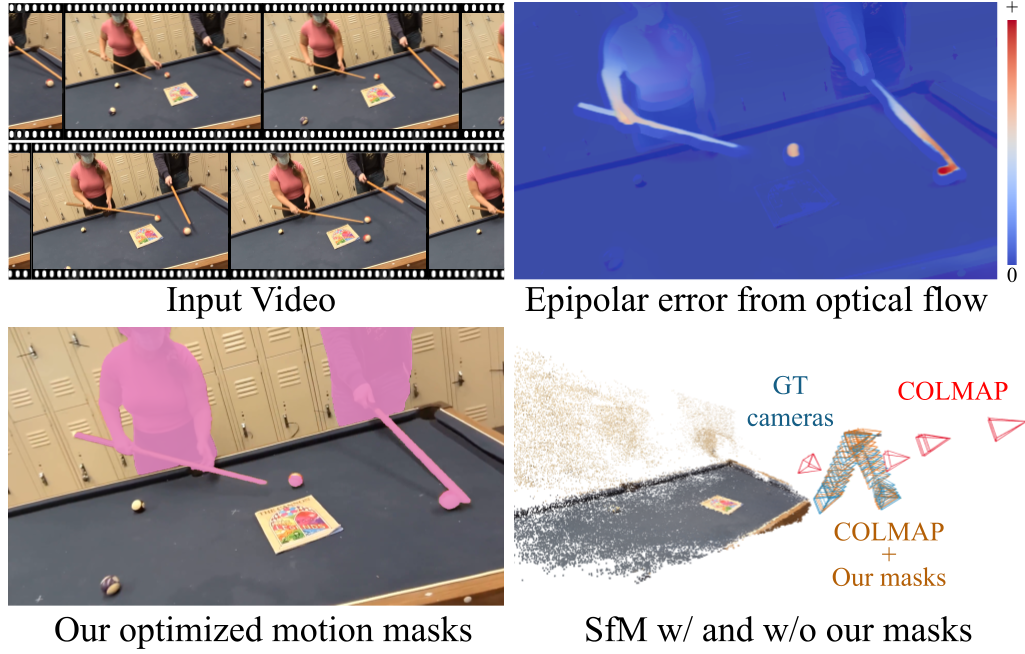}
\caption{
We introduce a zero-shot motion segmentation method for video based on cues from epipolar geometry (top right) and optical flow. Our predicted masks (bottom left) can help improve SfM camera calibration on highly dynamic scenes (bottom right).
}
\label{fig:teaser}
\end{figure}

Despite its potential application, the \textit{motion} segmentation task has been somewhat under-explored compared to image and video segmentation.
There exist supervised methods~\cite{Xie2022oclr, Lamdouar2021simo, Zhang2024monst3r}, but given the scarcity of real-world annotated data, most such techniques rely heavily on synthetic training data.
There are unsupervised motion segmentation methods~\cite{Koh2017arp, Meunier2023stfs, meunier2023emfs, Yang2021dystab, Yang2021mg}, but these do not exploit 3D geometric constraints, and tend to under-perform supervised methods.
Robust SfM pipelines for dynamic scenes~\cite{Zhao2022particlesfm, Zhang2022casualsam, Chen2024leapvo} exploit 3D geometric cues to identify problematic correspondences on dynamic objects but provide sparse masks for moving objects rather than densely segmenting entire objects.

This paper introduces a simple yet remarkably effective iterative approach to motion segmentation (see \cref{fig:teaser}).  It combines  optical flow and 3D geometric cues, along with a rich feature space from an off-the-shelf segmentation foundation model to facilitate the inference of coherent moving object masks.
In particular, given camera pose estimates, epipolar constraints can be used to predict which flow correspondences are inconsistent with the estimated camera poses~\cite{Zhang2022casualsam,Chen2024leapvo,Liu2023rodynrf}.
These sparse outliers then anchor the inference of segmentation masks for moving objects as a form of clustering in the feature space of a foundation model pre-trained on image and video segmentation tasks.
By repeating these steps, we iteratively refine our camera pose estimates, the detection of correspondence outliers, and the motion segmentation masks.

The resulting method, dubbed \methodshortname, outperforms both synthetically supervised and unsupervised methods on motion segmentation benchmarks~(DAVIS16~\cite{Perazzi2016davis}, SegTrackv2~\cite{Li2013segtrackv2} and FBMS59 \cite{Ochs2014fbms59}).
We show the approach also significantly outperforms the SoTA robust SfM methods for estimating camera pose on dynamic scene benchmarks (e.g., MPI Sintel~\cite{Butler2012sintel}). 
To assess performance of the SfM estimates beyond synthetic benchmarks, we collected a dataset of real scenes with ground truth camera motion.
On this new dataset our new SfM pipeline, leveraging \methodshortname to identify and discard moving objects, outperforms the previous SoTA methods by a substantial margin.

\section{Related work}
\label{sec:related}
\vspace{-0.05cm}

The segmentation of moving objects relative to a world coordinate frame (vs an egocentric frame), a.k.a.\ {\em motion segmentation}, is a long standing problem \citep{wu1993gradient, odobez1995mrf, wang1994representing}.
Early works \cite{odobez1995mrf, black1991robust, ochs2011object} relied on robust estimation and hierarchical layered representations to jointly model the motion field and object masks.
Supervised learning methods have also been used for geometric and semantic motion segmentation \citep{bideau2018best}.
While deep learning with large scale data has recently revolutionized many aspects of image and video understanding \citep{Ravi2024samv2, saxena2023ddvm, dpt2021}, progress in motion segmentation has been limited by the paucity of supervised training data, motivating unsupervised approaches.

\subsection{Flow based motion segmentation}
\vspace{-0.05cm}

With advances in optical flow (e.g., \texttt{RAFT}~\cite{raft2020}), flow-based segmentation methods have emerged~\citep{meunier2023emfs, Meunier2023stfs, Yang2021dystab, Yang2021mg}.
\citet{meunier2023emfs} uses an iterative EM procedure to segment the optical flow field into layers. 
\texttt{STM}~\cite{Meunier2023stfs} improves temporal consistency by using a spatio-temporal parametric motion model with a temporal consistency loss.
\texttt{Motion Grouping}~\cite{Yang2021mg} trains a self-supervised slot-attention auto-encoder to decompose flow fields into layers of background and foreground masks, iteratively aggregating regions with similar motion.
\texttt{DyStaB}~\cite{Yang2021dystab}, while not fully unsupervised, trains a segmentation network to minimize the mutual information between different segments of a flow field.
Similarly, \texttt{ARP}~\cite{Koh2017arp} uses both optical flow and RGB appearance of the frames to predict the motion masks.

Other methods leverage synthetic motion segmentation datasets~\citep{Lamdouar2021simo, Xie2022oclr}.
\texttt{SIMO}~\cite{Lamdouar2021simo} proposed large-scale synthetic data generation for training a transformer model for motion segmentation.
\texttt{OCLR-flo}~\cite{Xie2022oclr} takes a similar approach for videos with multiple objects and complex occlusions. To leverage appearance information (\texttt{OCLR-adap}) they additionally finetune DINO~\cite{dino} on their predicted masks (from flow only) and use it for mask propagation.

Similarly we use optical flow, but we further leverage epipolar geometry and features from a large semantic segmentation model to more robustly disambiguate between object motion and camera motion.

\subsection{Video object segmentation}
\vspace{-0.05cm}

Video object segmentation (VOS) aims to segment foreground objects in video regardless of motion.
VOS is easier to annotate with several public  datasets~\cite{Perazzi2016davis} however, `foreground' objects can be static and `background' objects can be dynamic which limits their usefulness for our task.
Unsupervised VOS methods~\cite{Wang2024videocutler, Lee2024gsa} can incorporate post-hoc motion identification techniques, potentially enabling motion segmentation.
Nevertheless, many lack the generality to segment uncommon dynamic objects, such as planted trees or shadows.
We take inspiration from unsupervised VOS in our use of semantic features, and from unsupervised motion segmentation in our use of optical flow, \textit{combining} them into a robust motion segmentation method applicable to in-the-wild SfM.
We furthermore make use of \texttt{SAMv2}~\cite{Ravi2024samv2} to improve resolution of our motion masks.

\subsection{Dynamic structure from motion}
\vspace{-0.05cm}

Widely used structure from motion methods such as \texttt{COLMAP}~\cite{schoenberger2016sfm} assume that scenes are largely static, and often fail on videos with dynamic objects.
\texttt{CasualSAM}~\cite{Zhang2022casualsam} addresses this by jointly optimizing for depth (using a learned pre-trained prior), camera poses and motion masks.
\texttt{ParticleSfM}~\cite{Zhao2022particlesfm} leverages off-the-shelf optical flow and monocular depth estimators to generate 3D tracks, and trains a 3D track motion classifier on synthetic data.
Only tracks corresponding to static parts of the scene are used for bundle adjustment.
\texttt{LEAP-VO}~\cite{Chen2024leapvo} similarly classifies tracks into static and dynamic elements, augments its inputs with features, has a refiner module, and a sliding window of tracks passed to a global bundle adjustment to determine camera poses.
DROID-SLAM~\cite{Teed2021droidslam} rejects moving correspondences by relying on the temporal consistency of objects in a GRU memory.
\texttt{DUSt3R}~\cite{Wang2023dust3r} is a novel pose inference technique using a patch based feed-forward network to predict global 3D coordinates.
\texttt{MonST3R}~\cite{Zhang2024monst3r} fine-tunes \texttt{DUSt3R} on scenes with dynamics.
Our method outperforms this SoTA methods on dynamic SfM~(\cref{fig:sintel}).
Dynamic objects also impose challenges for camera calibration in 3D reconstruction methods. \texttt{RoDynRF}~\cite{Liu2023rodynrf} is a 4D pose-free reconstruction method that optimizes cameras by removing scene elements with unreliable epipolar geometry using Sampson error~\cite{sampson82, sampson96} similar to us.

\section{Method}
\label{sec:method}
\vspace*{-0.1cm}

Given a video sequence of images $\{\image_t\}_{t=1}^{T}$, our goal is to estimate the corresponding pixel binary motion masks $\{\mask_t\}$,
\begin{equation}
    \mask_t^\uv = \indicator(\image_t^\uv \in \dynamics),
\end{equation}
where $\uv$ are 2D pixel coordinates, and $\dynamics$ is the set of pixels of dynamic objects.
We propose an iterative approach consisting of two key steps:
\begin{enumerate*} [label=(\arabic*)]
\item identify likely \textit{static}
pixels by considering optical flow between adjacent images in time, and using epipolar geometry to identify pixels in the scene whose movement can be explained \textit{solely} by changes in camera pose~(see \cref{sec:epipolar}); 
\item use these noisy labels together with features from a pre-trained video segmentation model to learn a classifier that produces higher-quality and temporally stable segmentation masks~(see \cref{sec:classifier}).
\end{enumerate*}
Iterating these steps refines the estimated epipolar geometry and, in turn, the predicted masks, resulting in further performance improvements~(see \cref{sec:iterative}).
Finally, to obtain higher-resolution segmentation masks, we again leverage the pre-trained video segmentation model~(see \cref{sec:spatial_ref}).

\begin{figure}
\centering
\includegraphics[width=\linewidth]{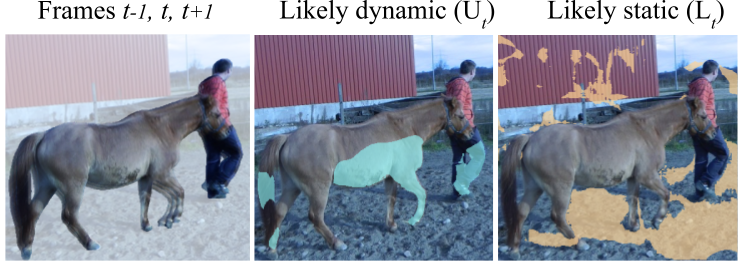}

\caption{\textbf{Epipolar matches (\cref{sec:epipolar}) --} 
$\upper_{t}$ and $\lowerr_{t}$ respectively capture the most \textit{likely} dynamic and static parts of the scene.
}
\label{fig:weakhints}
\end{figure}

\subsection{Weak epipolar supervision}
\label{sec:epipolar}
\vspace*{-0.05cm}
We start by pre-computing forward and backward optical flow fields $\flow_{t \rightarrow t+1}$ and $\flow_{t \rightarrow t-1}$ using \texttt{RAFT}~\cite{raft2020}.
Optical flow establishes a set of dense correspondences between two frames such that $\x'_{t'} = \x_t + \flow_{t \rightarrow t'}(\x_t)$.
To remove noisy correspondences (e.g. occlusion), we remove correspondences that do not pass a cycle consistency: where $\flow_{t' \rightarrow t}(\x'_{t'})$ does not return $\x'_{t'}$ to its original position $\x_t$.

\paragraph{Robust fundamental matrix estimation}
For a static scene, if pairwise correspondences $(\x_t, \x'_{t'})$ were pixel-perfect, we could employ the 7-point algorithm~\cite{Szeliski2011cv} to find the fundamental matrix $\fundamental_{tt'}$, and estimate the relative camera pose between the two frames.
To account for spurious correspondences caused by dynamic objects, we employ \texttt{RANSAC}~\cite{Fischler1981ransac} to robustly estimate $\fundamental_{tt'}$ with a Median of Squares consensus measure~\cite{Hampel1975lmeds}.

\paragraph{Scoring flow correspondences}
Having estimated the fundamental matrix $\fundamental_{tt'}$, we can use it to evaluate the quality of each flow correspondence $(\x_t, \x'_{t'})$ through epipolar geometry.
To this end we use the Sampson distance~\cite{sampson82, sampson96} as a linear approximation of the re-projection error:
\begin{equation}
    \sampson_{tt'}(\x_t) = \frac{|h(\x_t)^\top \fundamental_{tt'} h(\x'_{t'})|^2}{\|d(\fundamental_{tt'}  h(\x_t))\|_2^2 + \|d(\fundamental_{tt'} h(\x'_{t'}))\|_2^2},
    \label{eq:samson}
\end{equation}
where $h(.)$ is the homogeneous representation of a point, and $d(.)$ is the non-homogeneous representation.
Using \cref{eq:samson} we can obtain a binary mask of points in $\image_t$ that are \textit{likely} to be static ($\lowerr_{t}$) and dynamic ($\upper_{t}$) as:
\begin{align}
   \lowerr_{t} &= \indicator(\text{max}(\sampson_{t,t+1}, \sampson_{t,t-1} ) < \threshlow)
   \label{eq:static}
   \\
   \upper_{t} &=  \indicator(\text{max}(\sampson_{t,t+1}, \sampson_{t,t-1} ) > \threshup).   
   \label{eq:dynamic}
\end{align}
where, to account for differences in camera speed variations in scenes, we first compute the average L2 norm of flow per image $\avgflow$, and set the thresholds above as $\threshup = 2\cdot\avgflow$ and $\threshlow = 0.01 \cdot \avgflow$.

\begin{figure}
\centering
\includegraphics[width=\linewidth]{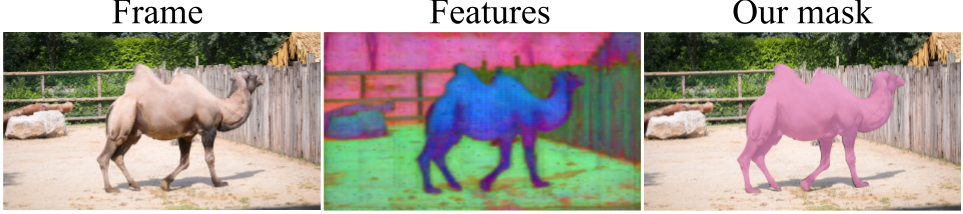}
\caption{\textbf{Feature-based classifier (\cref{sec:classifier}) --}
Feature space of foundation models show strong objectness prior as shown by the first three PCA components of the features. We leverage these features to train our classifier on sparse and noisy labels from epipolar supervision, generating coherent motion masks.
}
\vspace*{-0.1cm}
\label{fig:classifier}
\end{figure}

\subsection{Feature-based classifier}
\label{sec:classifier}
\vspace*{-0.05cm}

The masks from~\cref{eq:static,eq:dynamic} provide a \textit{sparse} and \textit{noisy} per \textit{image-pair} representation of the static/dynamic decomposition of a scene; see Fig.~\ref{fig:weakhints}.
Conversely, our objective is to produce a \textit{dense}, \textit{robust} and \textit{temporally consistent} pixel-by-pixel estimate of $\mask_t^\uv$.
To achieve this, we take advantage of the well-behaved feature space of the pre-trained SAMv2 video segmentation model~\cite{Ravi2024samv2}.
The intuition is that pixels corresponding to the \textit{same} object in a video should be close in feature space, as these models are specifically trained for object-level segmentation.
Within this feature space, we learn a lightweight classifier supervised with labels given by \cref{eq:static,eq:dynamic}; e.g., see Fig.\ \ref{fig:classifier}.
In particular, given image $\image_t$, we extract the corresponding feature map~$\feat_t$ from the last layer of the SAMv2 encoder and train a shallow multi-layer perceptron $\mlp_\param$ to classify the feature space via the loss~\cite{Sabour2024sls}:
\begin{align}
\mathcal{L}^t_\text{sup} 
= & \kappa(\text{max}(\upper_t - \mlp_\param(\feat_t), 0)) 
\nonumber
\\
&+ \kappa(\text{max}(\mlp_\param(\feat_t) - \overline{\lowerr}_t, 0)),
\end{align}
where $\overline{\lowerr}_t$ is the binary complement of ${\lowerr}_t$ and the use of $\text{max}$ ensures that $\mlp$ is only supervised at the pixel locations activated in $\upper_t$ or $\lowerr_t$.
We add a Geman-McClure robust kernel $\kappa$ with a fixed temperature to make our solution more robust to imperfections to the automatic selection of $\threshup$ and $\threshlow$ across sequences.
We also regularize the MLP weights via a Lipschitz regularizer $\loss_\text{reg}$ proposed by~\cite{Liu2022lipschitz}, and train across all time steps within a video sequence:
\begin{equation}
    \loss(\param) = \tfrac{1}{T} \sum_t \loss_\text{sup}^t(\param) + \loss_\text{reg}(\param).
\end{equation}
{Once trained, motion masks are found by thresholding:}
\begin{equation}
\mask_t = \indicator(\mlp_\param(\feat_t) > 0.5)
\label{eq:coarsemask}
\end{equation}

\paragraph{Dropping unreliable frames}
In our datasets, we observed extreme situations where foreground objects \textit{completely} occlude the scene.
This results in highly unreliable estimates of $\fundamental_{tt'}$, and therefore low-quality pseudo-labels $\upper$ and $\lowerr$ for supervising $\mlp_\param$.
We therefore drop frames whenever less than 50\% of pixels are marked as inliers, to boost performance; see the ablation in \Cref{sec:ablations}.

\begin{figure}
\centering
\includegraphics[width=\linewidth]{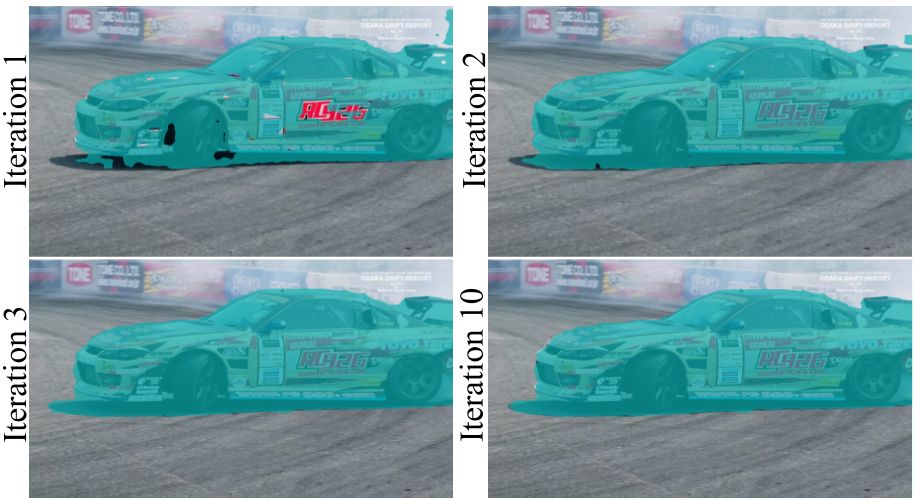}
\caption{\textbf{Iterative refinement (\cref{sec:iterative}) --} 
Repeated fundamental matrix estimation and motion prediction improves estimated camera pose and masks, often converging after 2 iterations.
}
\label{fig:iterative}
\end{figure}

\subsection{Iterative refinement}
\label{sec:iterative}
\vspace*{-0.05cm}

Once our classifier $\mlp_\param$ is trained, the predicted masks better approximate the moving objects than $\upper$ and $\lowerr$ as $\mlp_\param$ also considers semantic appearance, rather than just epipolar geometry.
Our iterative refinement process entails the use of $\mlp_\param$ to remove bad correspondences, re-computing the fundamental matrices, yielding improved masks $\upper$ and $\lowerr$, and using the updated masks to {fine-tune $\mlp_\param$ from the previous step}. 
We find that two iterations is optimal, after which  performance saturates; {see \cref{fig:iterative} and our ablation in \cref{sec:ablations}}.

\begin{figure}
\centering
\includegraphics[width=\linewidth]{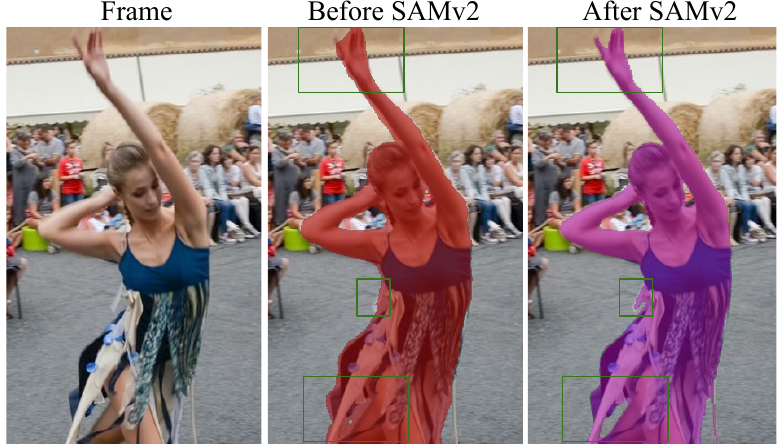}

\caption{\textbf{Final refinement (\cref{sec:spatial_ref}) --}
With SAMv2 we improve the fine-grained details in the mask.
In particular, note the finer details around the fingers and the dress frills.
}
\vspace*{-0.2cm}
\label{fig:spatial_ref}
\end{figure}

\subsection{Final refinement}
\label{sec:spatial_ref}
\vspace*{-0.05cm}

As the feature maps $\feat_t$ are of relatively low-resolution, our segmentation masks are initially coarse.
However, SAMv2 allows for coarse masks, points and boxes to be specified as input for any frame, and outputs temporally consistent fine-grained video segmentation masks.
We exploit this capability, and provide our coarse masks from~\cref{eq:coarsemask} to infer higher-resolution masks; see \cref{fig:spatial_ref} and the ablation in \cref{sec:ablations}.
\begin{figure*}[t]
\centering
\setlength\tabcolsep{7pt}
\includegraphics[width=\linewidth]{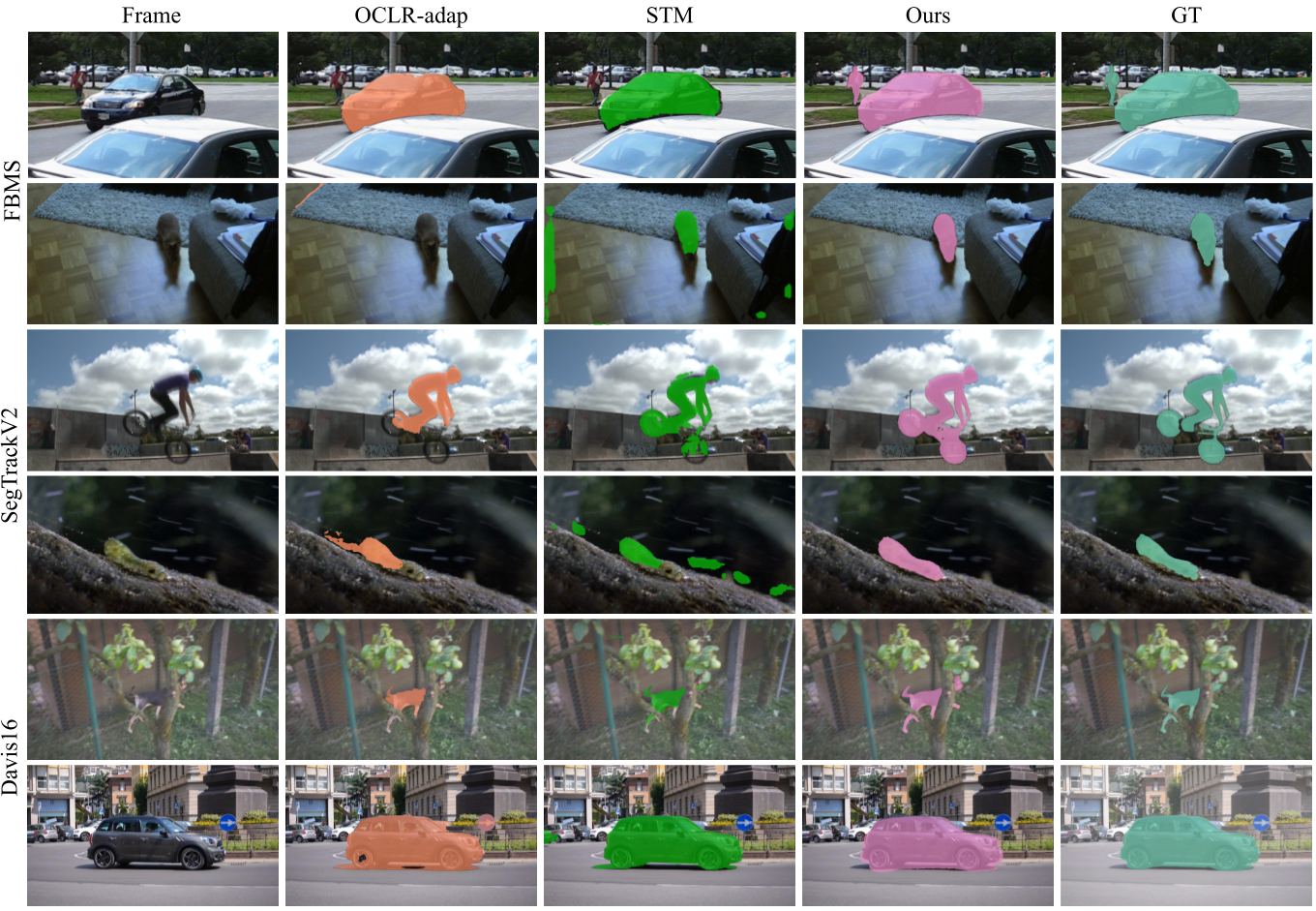}\\
\vspace{-.2cm}
\setlength{\tabcolsep}{24pt}
\resizebox{\linewidth}{!}{
\begin{tabular}{cccccc}
\toprule
  Training    &         Method                  &   DAVIS2016                     & SegTrackV2                   & FBMS59                  \\[.2em]
\midrule 
\multirow[c]{3}{*}{ \makecell{\shortstack{Unsupervised \\ on DAVIS2016}}}  & Motion Grouping~\cite{Yang2021mg}        & 68.3                        & 58.6                        & 53.1                        \\

&  EM ~\cite{meunier2023emfs}                               &         69.3                        & 55.5                        & 57.8                        \\
&  STM ~\cite{Meunier2023stfs}                             &         {\color[HTML]{000000} 73.2} & {\color[HTML]{000000} 55.0} & {\color[HTML]{000000} 59.2} \\[.2em]
\midrule
None & \textbf{ Ours (\methodshortname)}                             &        \textbf{77.3}               & \textbf{67.7}               & \textbf{75.5} 
\\%[.2em]
\midrule
 \multirow[c]{3}{*}{{\color[HTML]{9B9B9B} \makecell{\shortstack{Supervised \\ on Synthetic}} }  }  & {\color[HTML]{9B9B9B} SIMO} ~\cite{Lamdouar2021simo}      & {\color[HTML]{9B9B9B} 67.8}                        & {\color[HTML]{9B9B9B} 62.2}                        & {\color[HTML]{9B9B9B} - }                          \\ 
&  {\color[HTML]{9B9B9B} OCLR-flo}~\cite{Xie2022oclr}  &  {\color[HTML]{9B9B9B} 72.1} & {\color[HTML]{9B9B9B} 67.6} & {\color[HTML]{9B9B9B} 70.0} \\ 
&  {\color[HTML]{9B9B9B} OCLR-adap}~\cite{Xie2022oclr} &   {\color[HTML]{9B9B9B} 80.9} & {\color[HTML]{9B9B9B} 72.3} & {\color[HTML]{9B9B9B} 69.8} \\
\bottomrule
\end{tabular}

}%
\captionof{figure}{
\textbf{Motion segmentation} -- results on \texttt{DAVIS2016}~\cite{Perazzi2016davis}, \texttt{SegTrackV2}~\cite{Li2013segtrackv2} and \texttt{FBMS59}~\cite{Ochs2014fbms59} shows qualitative and quantitative (IoU $\uparrow$) improvement over unsupervised methods, and competitive results with supervised methods trained on synthetic data.
{While both \methodshortname and the unsupervised baselines do not require annotations, our method does not require any training whatsoever and can be applied zero-shot to a test video.}
}
\vspace*{-0.3cm}
\label{fig:mseg}
\end{figure*}
\section{Results}
\label{sec:results}
\vspace*{-0.1cm}

In \cref{sec:motionseg_eval}, we evaluate our method for motion segmentation, comparing its performance against baselines.
In \cref{sec:sfm_eval}, we evaluate our method's ability to improve camera estimation with different SfM methods.
Finally, in \cref{sec:ablations} we ablate different aspects of our method.

\paragraph{Implementation details}
We use  RAFT~\cite{raft2020} to compute optical flow between adjacent frames.
SAMv2 \cite{Ravi2024samv2} features from the last layer of the image encoder are used for training $\mlp_\param$, which has 1 hidden layer with 8 neurons.
It is trained for 25 epochs per refinement iteration using the Adam~\cite{Diederik2014adam} optimizer with a learning rate of $0.02$. 
We fix the Geman-McClure temperature parameter $\tau^2 {=} 0.01$. 
We use the SAMv2 video processor to post-process the masks. 
Experiments are conducted on a single NVIDIA A100 GPU.

\subsection{Motion segmentation}
\label{sec:motionseg_eval}
\vspace*{-0.1cm}

The task of motion segmentation has few dedicated benchmarks in either the supervised or unsupervised settings. 
Prior works, such as~\citet{meunier2023emfs} and \citet{Yang2021mg}, typically use a subset of established VOS benchmarks to evaluate motion masks.
As in prior works, we report the Jaccard score (i.e. Intersection over Union~(IoU)).

\begin{figure}[t]
\centering
\setlength\tabcolsep{3pt}
\includegraphics[width=\linewidth]{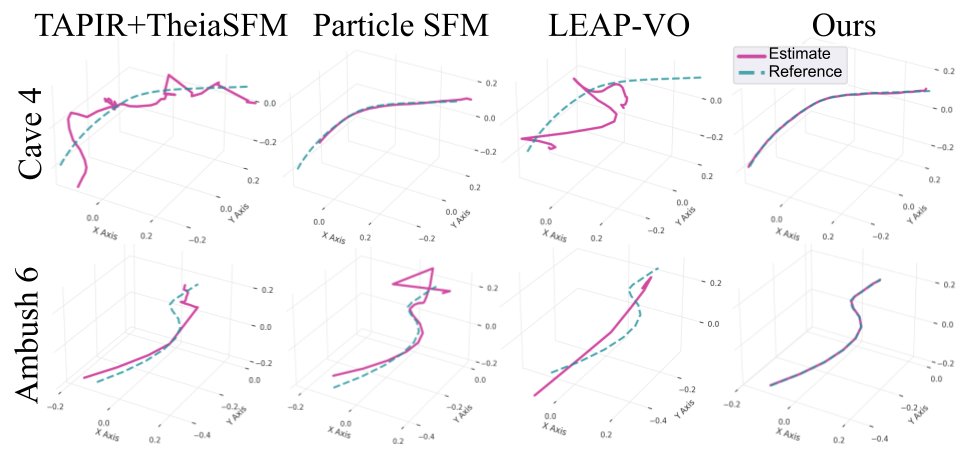}\\
\vspace*{0.1cm}
\resizebox{\linewidth}{!}{
\begin{tabular}{c|cccccc}
          & CasualSAM & MonST3R & LEAP-VO        & ParticleSFM & \cite{Doersch2023tapir}+\cite{Sweeney2015theiasfm} &
         \cite{Doersch2023tapir}+\cite{Sweeney2015theiasfm}+Ours           \\
          \hline
ATE $\downarrow$  & 0.141      & 0.108   & \textbf{0.089} & 0.129        & 0.132          & \underline{0.093}    \\
RPE-T $\downarrow$& 0.035      & 0.042   & 0.066          & \underline{0.031}  & 0.045          & \textbf{0.026} \\
RPE-R $\downarrow$& 0.615      & 0.732   & 1.25           & \underline{0.535}  & 1.175          & \textbf{0.217}
\end{tabular}
}%
\captionof{figure}{
\textbf{Camera calibration (MPI Sintel)} -- 
Bundle adjustment with \texttt{TheiaSFM}~\cite{Sweeney2015theiasfm} on dense correspondences from \texttt{TAPIR}~\cite{Doersch2023tapir} is
improved when correspondences are masked with \methodshortname. 
}
\vspace*{-0.4cm}
\label{fig:sintel}
\end{figure}

\paragraph{Datasets}
We use \texttt{DAVIS2016} \cite{Perazzi2016davis}, a dataset of 50 videos with moving objects and cameras.
Following ~\cite{Meunier2023stfs, Xie2022oclr}, we evaluate on the 20 validation sequences.
\texttt{DAVIS2016} only annotates a single prominent object for each frame, even though some sequences have unannotated moving background objects. 
Our method is designed to detect \emph{any} movement in the scene aside from ego-motion but, due to the lack of ground truth annotations, we resort to comparing the full predicted motion mask with the partial annotated mask as given in the dataset.
We also use \texttt{SegTrackV2}~\cite{Li2013segtrackv2} and \texttt{FBMS59}~\cite{Ochs2014fbms59}, which respectively have 14 and 59 videos with moving objects and a moving camera.
We use the 30 sequences annotated as the evaluation set in \texttt{FBMS59} to be consistent with baseline methods.
We use the same evaluation setup as \citet{meunier2023emfs} and in the case of several moving objects, group
them all into a single foreground mask for evaluation.

\paragraph{Baselines}
We compare against state-of-the-art methods for unsupervised motion segmentation, including \texttt{Motion Grouping}~\cite{Yang2021mg}, \texttt{EM}~\cite{meunier2023emfs}, and \texttt{STM} \cite{Meunier2023stfs}.
These methods infer masks for any object with a distinct motion by reasoning about the optical flow.
They do not necessarily only mask objects that have motion relative to a rigid world frame, hence the results from these methods are usually post processed to keep only the mask that has the highest similarity to the ground truth mask~\cite{Meunier2023stfs}.
We also include \texttt{OCLR-flo}~\cite{Xie2022oclr} and \texttt{SIMO}~\cite{Lamdouar2021simo} as more competitive methods, although both these  have been trained in a supervised manner on synthetic datasets.
\texttt{OCLR-adap}~\cite{Xie2022oclr} is their test-time adaptation variant, which fine-tunes on test video.
\texttt{ARP}~\cite{Koh2017arp} and \texttt{DyStab-Dyn}~\cite{Yang2021dystab} are excluded as baselines due to a lack of publicly available source code.

\paragraph{Quantitative analysis -- \Cref{fig:mseg}}
Our zero-shot method significantly outperforms existing unsupervised motion segmentation methods on \textit{all three} VOS benchmarks, and shows competitive performance with synthetically supervised methods.
Note the significant improvement on \texttt{FBMS59}.
This is likely because \texttt{FBMS59}, as a motion segmentation dataset, has \textit{all} moving objects properly annotated, unlike \texttt{DAVIS16} and \texttt{SegTrackV2}.

\paragraph{Qualitative analysis -- \Cref{fig:mseg}}
The qualitative results illustrate several aspects of our method:
\begin{enumerate*} [label=(Row \arabic*)]
\item shows a challenging example where we correctly segment a slow moving pedestrian in the background which is typically missed the prior works;
\item baselines either completely miss or over-segment frames where a dynamic object is \textit{momentarily} static;
\item our method is robust to motion blur;
\item most prior work fails to segment the full camouf lagedobject without including some of the background;
\item part of the animal is occluded behind the tree trunk. As such the animal includes several segments. While prior work miss one or more segments of the animal we are able to segment it correctly.
\item \texttt{OCLR-adap} masks the static sign with similar depth to the car.
\end{enumerate*}

\subsection{Applications to SfM}
\label{sec:sfm_eval}
\vspace*{-0.1cm}
A direct application of motion segmentation is the improvement of camera trajectory estimation for video with both dynamic objects and dynamic cameras.

\begin{figure}[t]
\centering
\setlength{\tabcolsep}{15pt}
\includegraphics[width=\linewidth]{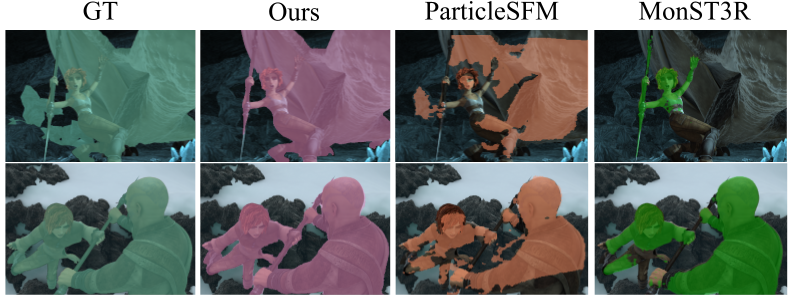}\\
\resizebox{\linewidth}{!}{
\begin{tabular}{c|ccc}
Dataset &  \textbf{Ours (\methodshortname)} & ParticleSFM &  MonST3R \\ \hline
Sintel IoU $\uparrow$ &  \textbf{69.6}      & 41.7     & 60.6 \\
\end{tabular}
}%
\captionof{figure}{
\textbf{Segmentation masks (MPI Sintel)} --
Average IoU of motion mask for widely used SfM methods for dynamic scenes. 
}
\label{fig:sintel-mask}
\vspace*{-0.5cm}
\end{figure}

\paragraph{Datasets}
In-the-wild SfM datasets are rare, especially when one requires ground truth camera trajectories, and most video sequences with ground truth cameras for the Structure from Motion~(SFM) task are from stationary scenes.
A solution to this problem has been to use synthetic datasets like \texttt{MPI Sintel}~\cite{Butler2012sintel} to evaluate camera pose estimation, as it offers ground-truth camera annotations of $23$ synthetic sequences of dynamic scenes.
Although this is a good starting point for performing evaluations on this task, \texttt{MPI Sintel} contains simple camera movements, the appearance is far from realistic, the scenes are generally textureless, and minimal 3D structure exists once the dynamic objects are removed.
Hence, we propose a new \textit{casual motion dataset} in Sec.~\ref{sec:casdata}, with $8$ real-world scene captured with a robotic arm, all with ground-truth camera annotations. 
Accordingly, we also evaluate baselines and our method on this dataset.
For both test datasets we report the absolute camera trajectory error (ATE) and the translation/rotation part of relative pose error (RPE-T and RPE-R). ATE and RPE-T are reported in meters, and in the same scale as ground truth trajectories. RPE-R is in degrees.

\begin{figure*}[t]
\centering
\setlength\tabcolsep{5pt}
\includegraphics[width=\linewidth]{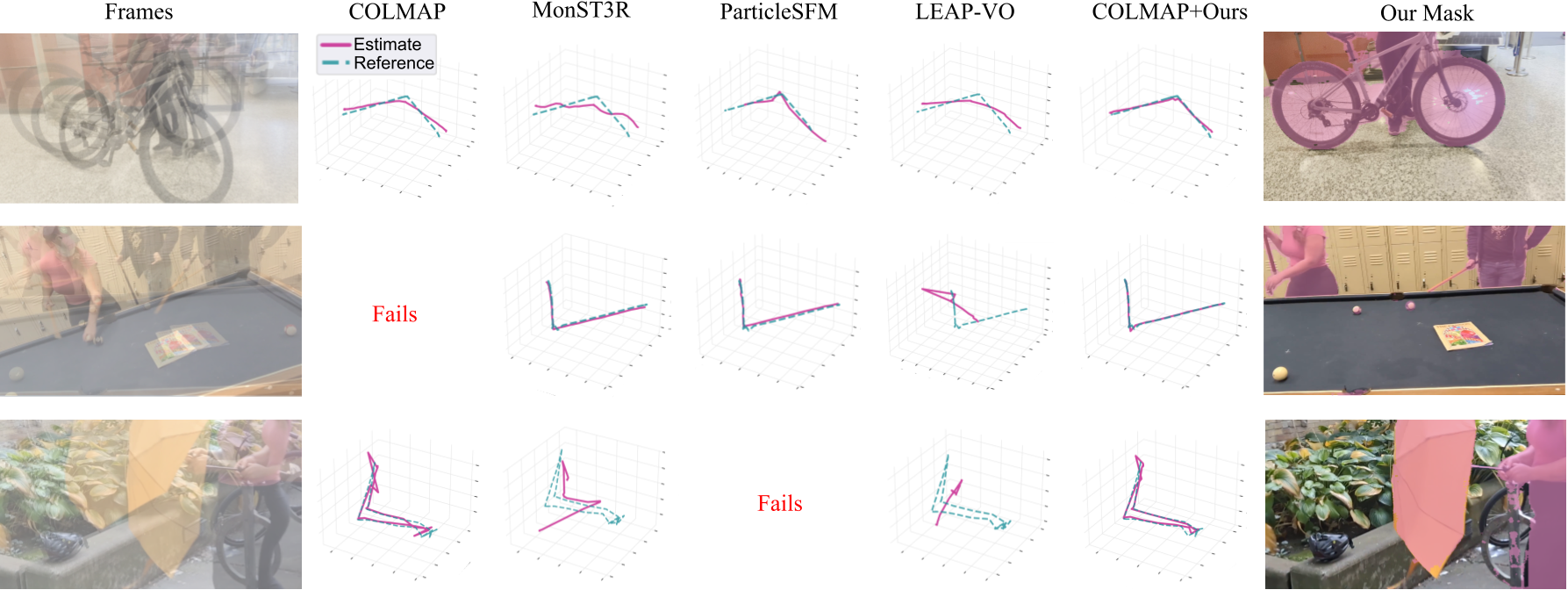}\\
\vspace{0.1cm}
\setlength{\tabcolsep}{14pt}
\resizebox{\linewidth}{!}{
\begin{tabular}{cccccccccc}
\toprule
\multirow{2}{*}{Method} & \multicolumn{3}{c}{All (8 scenes)}                      & \multicolumn{3}{c}{COLMAP Subset (7 scenes)} & \multicolumn{3}{c}{ParticleSFM Subset (5 scenes)}             \\

                        & ATE  $\downarrow$         & RPE-T $\downarrow$      & RPE-R   $\downarrow$    & ATE       $\downarrow$    & RPE-T $\downarrow$      & RPE-R $\downarrow$  & ATE   $\downarrow$        & RPE-T  $\downarrow$     & RPE-R  $\downarrow$   \\ \midrule
COLMAP                  & -             & -             & -             & 0.70          & 0.63          & 2.17     & -          & -          & -      \\
Monst3r                 & 1.09          & 0.72          & 2.25          & 1.20          & 0.71          & 2.39      & 1.05          & 0.71          & 2.20      \\
ParticleSFM             & -          & -          & -          & -          & -          & -      & 0.87          & 0.61          & 2.00      \\
LEAP-VO                 & 2.10          & 0.77          & 3.86          & 1.96          & 0.72          & 3.25    & 2.50          & 0.87          & 4.69        \\
\textbf{Ours (\methodshortname)}                    & \textbf{0.41} & \textbf{0.48} & \textbf{1.60} & \textbf{0.40} & \textbf{0.45} & \textbf{1.62} & \textbf{0.39}          & \textbf{0.56}          & \textbf{1.98} \\
\bottomrule
\end{tabular}
}%
\captionof{figure}{
\textbf{Evaluation on our 
dataset} -- 
SfM results on our challenging real-world dataset, including multiple human actors, high occlusion and fast camera movements, shows that our masking method can improve a simple classic SfM method like COLMAP~\cite{schoenberger2016sfm} to outperform the previous SOTA methods on dynamic scene SfM. 
We also report on subsets of scenes, as some methods fail to converge on some scenes.
}
\label{fig:ourdataset}
\vspace*{-0.3cm}
\end{figure*}

\vspace*{-0.1cm}
\paragraph{Baselines}
We evaluate \methodshortname against \texttt{CasualSAM}~\cite{Zhang2022casualsam}, \texttt{MonST3R}~\cite{Zhang2024monst3r}, \texttt{LEAP-VO}~\cite{Chen2024leapvo}, and \texttt{ParticleSFM}~\cite{Zhao2022particlesfm}, all designed to handle the dynamic objects by masking their  correspondences
or by estimating static 3D pointmaps and 2D dynamic masks.
\texttt{ParticleSFM} classifies dense tracks initialized with flow as either dynamic or static and then uses the filtered tracks for global bundle adjustment~\cite{Wilson2014odsfm, Sweeney2015theiasfm}.
\texttt{LEAP-VO} has a similar strategy of finding reliable point tracks and classifying them as static versus moving or invisible trajectories.
\texttt{CasualSAM} optimizes camera parameters together with motion and depth maps and uses the predicted motion field to inversely weigh the depth and flow reconstruction optimization objective, as a way of lowering the effect of dynamic pixels in the estimation.
Finally, \texttt{MonST3R} is a learning-based method
in which 3D points are directly regressed from pairs of images.
\texttt{MonST3R} is fine-tuned on dynamic scenes to provide more accurate pointmaps and predicts 2D motion masks using the regressed pointmaps. We also compare to naive baselines of \texttt{COLMAP}~\cite{schoenberger2016mvs, schoenberger2016sfm} with unmasked data, and global bundle adjustment \cite{Sweeney2015theiasfm} applied to unmasked dense tracks from~\cite{Doersch2023tapir}. 

\vspace*{-0.1cm}
\paragraph{Implementation details}
Since \methodshortname is not a full SfM framework, we need to pair our method with an SfM method. Typically, \texttt{COLMAP} is a go-to method for SfM and accepts masks as input to apply on extracted features.
However in case of the MPI Sintel~\cite{Butler2012sintel} dataset, due to its synthetic nature, the remaining scene after masking is particularly plain and featureless, leading to ineffective feature extraction with the classical \texttt{SIFT}~\cite{Lowe199sift} feature extraction used in \texttt{COLMAP}.
We therefore further run \texttt{TAPIR}~\cite{Doersch2023tapir} on the video to extract dense tracks as feature correspondences.
{We then use the same global bundle adjustment method as \texttt{ParticleSFM}~\cite{Zhao2022particlesfm} that uses \texttt{1DSfM}~\cite{Wilson2014odsfm} in the \texttt{TheiaSFM}~\cite{Sweeney2015theiasfm} library.
\texttt{ParticleSFM} shows that \texttt{TheiaSFM} is effective in finding dense correspondences.}

\vspace*{-0.05cm}
\paragraph{Analysis (camera calibration) -- \Cref{fig:sintel}}
We evaluate on \texttt{MPI Sintel}~\cite{Butler2012sintel} with the same protocol as \texttt{ParticleSFM}~\cite{Zhao2022particlesfm} that removes invalid sequences~(e.g. static cameras), resulting in 14 sequences.
We show significant improvement in camera trajectory prediction in terms of RPE-R against state-of-the-art methods. Furthermore, qualitative visualizations of camera trajectories also demonstrate  superior performance.

\vspace*{-0.05cm}
\paragraph{Analysis (motion masking) -- \Cref{fig:sintel-mask}}
The comparison of our masks on the \texttt{MPI Sintel} dataset to the other SoTA dynamic SfM baselines, which use motion masking in their method, verifies that our method has superior motion segmentation performance on these scenes.

\subsection{New ``Casual Motion'' dataset -- Figure \ref{fig:ourdataset}}
\label{sec:casdata}
\vspace*{-0.05cm}
Estimating the camera trajectory of video frames with both camera and object motion is challenging. When an SfM model fails on a scene it is ambiguous whether the failure is due to the foreground motion, or to the lack of background details and depth variation. Therefore, the best current benchmarks aiming at the task of in-the-wild SfM, replicating casual captures, are synthetic, and do not capture the nuances of the real world.
To this end, we designed a capture method using a mobile robotic arm that traces a repeatable trajectory multiple times for in-the-wild SfM evaluation.
We first capture a clean sequence which is free of any moving objects, passing it to \texttt{COLMAP} to compute the ground-truth camera trajectory.
We then recapture the same trajectory in a  \textit{cluttered} setup, which includes moving objects and people in the scene.
This dataset includes a range of moving object, occlusion rates, rigid and deformable bodies, and indoor/outdoor scenes.
Our new dataset includes eight scenes, each with 40 to 50 frames, and two modalities (clean or cluttered).
We capture videos with the front camera of an iPhone12 attached to a Franka Emika Panda~\cite{Franka2024panda} robotic arm controlled with the Polymetis library \cite{polymetis}.

\vspace*{-0.05cm}
\paragraph{Analysis}
We evaluate our motion segmentation method paired with \texttt{COLMAP} on our dataset, and compare to \texttt{LEAP-VO}, \texttt{MonST3R} and \texttt{COLMAP}.
The results in \cref{fig:ourdataset} show that our method outperforms all baselines both qualitatively and {quantitatively}.
Notably, \texttt{COLMAP} fails completely on 1 out of 8 scenes and \texttt{ParticleSFM} fails on 3 out of 8 scenes.
We include the average results for each of these two methods on the separate subset where they do not have any failures (see supplementary for a detailed per scene evaluation).
We empirically observe that the use of dense tracks as opposed to \texttt{SIFT} features does not lead to significant improvement in this dataset, as the static part of the scene has enough texture and geometry.
This is contrary to the typical necessity of making use of tracks on synthetic datasets, such as \texttt{MPI~Sintel}, further emphasizing that evaluating SfM methods only on synthetic scenes may be insufficient. Note ground truth camera trajectory scale, and hence ATE and RPE-T, are in the arbitrary scale of the \texttt{COLMAP} solution for the clean Casual Motion dataset.

\begin{figure}[t]
\centering
\setlength\tabcolsep{7pt}
\includegraphics[width=\linewidth]{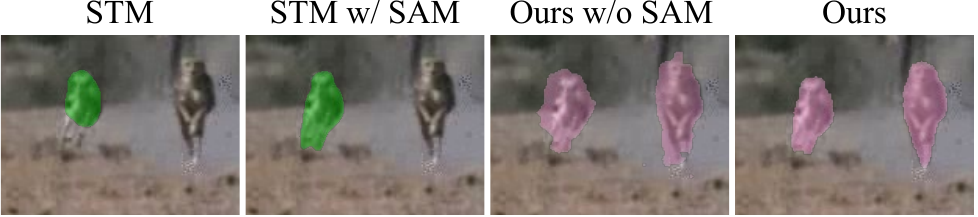} \\
\resizebox{\linewidth}{!}{
\begin{tabular}{c|ccccc}
IoU Avg. $\uparrow$ &   STM   &  \makecell{STM w/ \\ \texttt{SAMv2}}  &   \makecell{Ours  w/o \\ \texttt{SAMv2}} & \makecell{Ours} & \makecell{Ours w/ \\ \texttt{SD}} \\
\hline
\makecell{Track Seg v2\\ (Subset)} & 53.82 & 72.20 & 68.68 & \textbf{73.08} & 71.53  \\                                           
\end{tabular}
}%
\captionof{figure}{
\textbf{Ablation (\texttt{SAMv2} features and post processing)} -- \texttt{SAMv2} feature space and \texttt{Stable Diffusion} feature space show similar effectiveness. \texttt{SAMv2} post processing improves performance for both our method {and~\texttt{STM}~\cite{Meunier2023stfs}}.
}
\label{fig:ablation}
\end{figure}

\begin{figure}[t]
\centering
\setlength\tabcolsep{7pt}
\includegraphics[width=\linewidth]{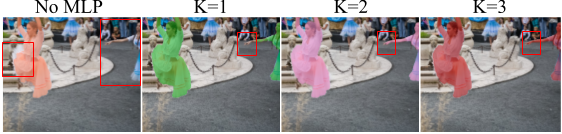}\\
\vspace*{0.1cm}
\resizebox{\linewidth}{!}{
\begin{tabular}{@{}cccccc|c@{}}
$\threshlow$($\avgflow$) & $\threshup$($\avgflow$) & MLP Size        & $\steps$ Iterations &  Kernel $\kappa$ & Drop  & DAVIS16  IoU $\uparrow$\\
\midrule
$2.0$ & $0.01$  &  --    & --   & \cmark &\cmark       &  70.9          \\
$2.0$ & $0.01$  &  8     & 1    &\cmark        &\cmark       &  78.1          \\
$2.0$ & $0.01$  &  8     & 2    &\cmark        &\cmark       &  \textbf{79.0} \\
$2.0$ & $0.01$  &  8     & 3    &\cmark        &\cmark       &  \textbf{79.0} \\ 
$2.0$ & $0.01$  &  8    & 2    &\xmark        &\cmark       &  78.3          \\
$2.0$ & $0.01$  &  8    & 2    &\cmark        &\xmark       &   78.5         \\
$1.0$ & $0.10$  &  8     & 2    &\cmark        &\cmark       &  77.8          \\
$2.0$ & $0.01$  &  32    & 2    &\cmark        &\cmark       &  75.8          \\
\end{tabular}
}%
\captionof{figure}{
\textbf{Ablation (classifier)} --
Processing outlier masks $\upper_t$ directly with \texttt{SAMv2} (no MLP) gives poor results.
Extra refinement iterations help, but the performance quickly saturates.
Higher capacity causes over-fitting and degradation of results.
Using a robust kernel and dropping unreliable frames do
yield some gains.
}
\vspace{-0.5cm}
\label{fig:mlpablate}
\end{figure}

\subsection{Ablations}
\label{sec:ablations}
\vspace*{-0.05cm}
We ablate the usage of \texttt{SAMv2}~\cite{Ravi2024samv2}, both for input features~(\cref{sec:classifier}) and our post-processing refinement step~(\cref{sec:spatial_ref}), on a subset of 13 scenes from \texttt{SegTrackV2}. 
We omit the `monkey' scene as it is a fully dynamic scene with no static pixels.
\cref{fig:ablation} shows that using off-the-shelf features from a fully unsupervised \texttt{Stable Diffusion}~\cite{Rombach2021high} model instead of \texttt{SAMv2} features has little to no impact in our motion segmentation task.
Post-processing with \texttt{SAMv2} to improve the mask boundaries however, improves the results by about $4\%$.
We further investigate our refinement step evaluating \texttt{SAMv2} refined \texttt{STM}~\cite{meunier2023emfs} masks, observing similar gains.

We ablate MLP classifier design choices on the 50 scenes from \texttt{DAVIS2016} in \cref{fig:mlpablate}.
Applying \texttt{SAMv2} refinement directly on the weak epipolar supervisory signal $\upper_t$ significantly decreases the average IoU of the predicted masks. 
Iterative refinement improves the results marginally in terms of quantitative measurements, and significantly in terms of qualitative results, but saturates quickly after $\steps=2$ iterations.
We also experiment with the choice of MLP size, threshold for upper/lower bounds, using Geman-McClure kernel and the effect of dropping unreliable frames.
\begin{figure}[t]
\centering
\includegraphics[width=\linewidth]{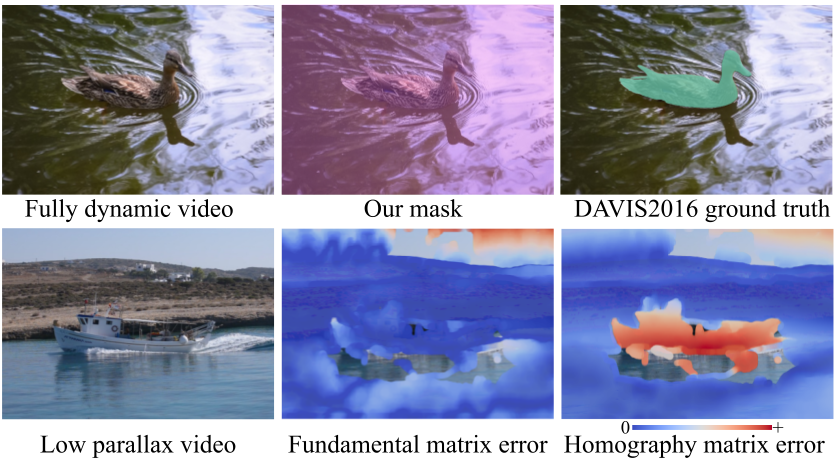}
\caption{
\textbf{Limitations} --
{(top) Our mask may cover the entire image when there are no visible static objects w.r.t.\ the world frame.
(bottom) The fundamental matrix solve fails on low parallax sequences, eg where a homography provides a better model.}
}
\label{fig:limitations}
\vspace*{-0.5cm}
\end{figure}

\section{Conclusions}
\vspace*{-0.1cm}
We introduce \methodshortname, a novel motion segmentation method with the goal of improving structure from motion for in-the-wild video.
We propose a novel iterative approach that combines epipolar constraints with semantic segmentation priors to predict accurate motion masks in challenging scenes.
Our results show that \methodshortname substantially outperforms existing unsupervised motion segmentation techniques on standard benchmarks.
We further evaluate the ability of our motion masks to improve dynamic SfM and demonstrate significant improvements in camera pose estimation.
Existing benchmarks are limited due to a lack of real-world video scenes with diverse motions and ground truth camera poses.
To enrich available benchmarks we collect and release a dataset (dubbed ``Casual Motion'') with challenging motion and groundtruth cameras.

Despite our proposed improvements, we observe certain limitations; see~\cref{fig:limitations}.
For example, in cases of low parallax, weak epipolar constraints struggle to effectively separate static vs dynamic regions.
In future work we will investigate the use of homographies, as suggested by~\cite{torr98}, in the low parallax scenes; see \cref{fig:limitations}~(bottom).
Another challenge occurs when frames lack \textit{any} static scene content, e.g., the video of a duck moving along the waves of a pond in~\cref{fig:limitations}~(top).
In this case our method masks out the full frame, which though technically correct, can be problematic for a downstream SfM method.
Finally, our use of off-the-shelf optical flow and semantic features means our method may be susceptible to their flaws but will similarly improve as they do.

\paragraph{Acknowledgements}
We thank Abhijit Kundu and Kevin Murphy for reviewing our manuscript and for their valuable feedback. 
We also would like to thank Karthik Mahadevan, Otman Benchekroun, Kinjal Parikh and Bogdan Pikula for their help in collecting data and Tovi Grossman for his generous support in our dataset collection effort.
Also thanks to Wesley Helmholtz for his help in the publication process.

\vspace*{0.2cm}

{
    \small
    \bibliographystyle{ieeenat_fullname}
    \bibliography{main}

\begin{thebibliography}{58}
\providecommand{\natexlab}[1]{#1}
\providecommand{\url}[1]{\texttt{#1}}
\expandafter\ifx\csname urlstyle\endcsname\relax
  \providecommand{\doi}[1]{doi: #1}\else
  \providecommand{\doi}{doi: \begingroup \urlstyle{rm}\Url}\fi

\bibitem[Bideau et~al.(2018)Bideau, RoyChowdhury, Menon, and
  Learned-Miller]{bideau2018best}
Pia Bideau, Aruni RoyChowdhury, Rakesh~R Menon, and Erik Learned-Miller.
\newblock The best of both worlds: Combining cnns and geometric constraints for
  hierarchical motion segmentation.
\newblock In \emph{CVPR}, 2018.

\bibitem[Black and Anandan(1991)]{black1991robust}
Michael~J Black and Padmanabhan Anandan.
\newblock Robust dynamic motion estimation over time.
\newblock In \emph{CVPR}, 1991.

\bibitem[Butler et~al.(2012)Butler, Wulff, Stanley, and
  Black]{Butler2012sintel}
D.~J. Butler, J. Wulff, G.~B. Stanley, and M.~J. Black.
\newblock A naturalistic open source movie for optical flow evaluation.
\newblock In \emph{ECCV}, 2012.

\bibitem[Caron et~al.(2021)Caron, Touvron, Misra, Jégou, Mairal, Bojanowski,
  and Joulin]{dino}
Mathilde Caron, Hugo Touvron, Ishan Misra, Hervé Jégou, Julien Mairal, Piotr
  Bojanowski, and Armand Joulin.
\newblock Emerging properties in self-supervised vision transformers.
\newblock \emph{ICCV}, 2021.

\bibitem[Chen et~al.(2024)Chen, Chen, Wang, and Pollefeys]{Chen2024leapvo}
Weirong Chen, Le Chen, Rui Wang, and Marc Pollefeys.
\newblock {LEAP}-{VO}: {Long}-term {Effective} {Any} {Point} {Tracking} for
  {Visual} {Odometry}.
\newblock In \emph{CVPR}, 2024.

\bibitem[Doersch et~al.(2023)Doersch, Yang, Vecerik, Gokay, Gupta, Aytar,
  Carreira, and Zisserman]{Doersch2023tapir}
Carl Doersch, Yi Yang, Mel Vecerik, Dilara Gokay, Ankush Gupta, Yusuf Aytar,
  Joao Carreira, and Andrew Zisserman.
\newblock {TAPIR}: {Tracking} {Any} {Point} with {Per}-{Frame} {Initialization}
  and {Temporal} {Refinement}.
\newblock In \emph{ICCV}, 2023.

\bibitem[Fischler and Bolles(1981)]{Fischler1981ransac}
Martin~A. Fischler and Robert~C. Bolles.
\newblock Random sample consensus: a paradigm for model fitting with
  applications to image analysis and automated cartography.
\newblock \emph{Commun. ACM}, 1981.

\bibitem[Franka Emika Panda()]{Franka2024panda}
Franka Emika Panda.
\newblock
  \url{https://download.franka.de/End-of-Life-Franka-Emika-Robot_EN.pdf}, 2024.

\bibitem[Hammer et~al.(2016)Hammer, Voit, and Beyerer]{Hammer2016msa}
Jan~Hendrik Hammer, Michael Voit, and Jürgen Beyerer.
\newblock Motion segmentation and appearance change detection based {2D} hand
  tracking.
\newblock In \emph{FUSION}, 2016.

\bibitem[Hampel(1975)]{Hampel1975lmeds}
F.~R. Hampel.
\newblock Beyond location parameters: Robust concepts and methods.
\newblock In \emph{Bulletin of the ISI}, 1975.

\bibitem[Kerbl et~al.(2023)Kerbl, Kopanas, Leimk{\"u}hler, and Drettakis]{3dgs}
Bernhard Kerbl, Georgios Kopanas, Thomas Leimk{\"u}hler, and George Drettakis.
\newblock 3d gaussian splatting for real-time radiance field rendering.
\newblock In \emph{ACM Transactions on Graphics}, 2023.

\bibitem[Kingma and Ba(2014)]{Diederik2014adam}
Diederik~P. Kingma and Jimmy Ba.
\newblock Adam: A method for stochastic optimization.
\newblock \emph{ICLR}, 2014.

\bibitem[Klappstein et~al.(2009)Klappstein, Vaudrey, Rabe, Wedel, and
  Klette]{Klappstein2009mos}
Jens Klappstein, Tobi Vaudrey, Clemens Rabe, Andreas Wedel, and Reinhard
  Klette.
\newblock Moving {Object} {Segmentation} {Using} {Optical} {Flow} and {Depth}
  {Information}.
\newblock In \emph{AIVT}, 2009.

\bibitem[Koh and Kim(2017)]{Koh2017arp}
Yeong~Jun Koh and Chang-su Kim.
\newblock Primary object segmentation in videos based on region augmentation
  and reduction.
\newblock In \emph{CVPR}, 2017.

\bibitem[Lamdouar et~al.(2021)Lamdouar, Xie, and Zisserman]{Lamdouar2021simo}
Hala Lamdouar, Weidi Xie, and Andrew Zisserman.
\newblock Segmenting {Invisible} {Moving} {Objects}.
\newblock In \emph{BMVC}, 2021.

\bibitem[Lee et~al.(2024)Lee, Cho, Lee, Park, Lee, and Lee]{Lee2024gsa}
Minhyeok Lee, Suhwan Cho, Dogyoon Lee, Chaewon Park, Jungho Lee, and Sangyoun
  Lee.
\newblock Guided {Slot} {Attention} for {Unsupervised} {Video} {Object}
  {Segmentation}.
\newblock In \emph{CVPR}, 2024.

\bibitem[Li et~al.(2013)Li, Kim, Humayun, Tsai, and Rehg]{Li2013segtrackv2}
Fuxin Li, Taeyoung Kim, Ahmad Humayun, David Tsai, and James~M. Rehg.
\newblock Video {Segmentation} by {Tracking} {Many} {Figure}-{Ground}
  {Segments}.
\newblock In \emph{ICCV}, 2013.

\bibitem[Lin et~al.(2021)Lin, Wang, Sutanto, Rai, and Meier]{polymetis}
Y. Lin, A.~S. Wang, G. Sutanto, A. Rai, and F. Meier.
\newblock Polymetis.
\newblock \url{https://facebookresearch.github.io/fairo/polymetis/}, 2021.

\bibitem[Liu et~al.(2022)Liu, Williams, Jacobson, Fidler, and
  Litany]{Liu2022lipschitz}
Hsueh-Ti~Derek Liu, Francis Williams, Alec Jacobson, Sanja Fidler, and Or
  Litany.
\newblock Learning smooth neural functions via lipschitz regularization.
\newblock In \emph{SIGGRAPH}, 2022.

\bibitem[Liu et~al.(2023)Liu, Gao, Meuleman, Tseng, Saraf, Kim, Chuang, Kopf,
  and Huang]{Liu2023rodynrf}
Yu-Lun Liu, Chen Gao, Andréas Meuleman, Hung-Yu Tseng, Ayush Saraf, Changil
  Kim, Yung-Yu Chuang, Johannes Kopf, and Jia-Bin Huang.
\newblock Robust {Dynamic} {Radiance} {Fields}.
\newblock In \emph{CVPR}, 2023.

\bibitem[Lowe(1999)]{Lowe199sift}
David~G. Lowe.
\newblock Object recognition from local scale-invariant features.
\newblock In \emph{CVPR}, 1999.

\bibitem[Luong and Faugeras.(1996)]{sampson96}
Quan-Tuan Luong and Olivier~D Faugeras.
\newblock The fundamental matrix: Theory, algorithms, and stability analysis.
\newblock In \emph{IJCV}, 1996.

\bibitem[Meunier and Bouthemy(2023)]{Meunier2023stfs}
Etienne Meunier and Patrick Bouthemy.
\newblock Unsupervised {Space}-{Time} {Network} for {Temporally}-{Consistent}
  {Segmentation} of {Multiple} {Motions}.
\newblock In \emph{CVPR}, 2023.

\bibitem[Meunier et~al.(2023)Meunier, Badoual, and Bouthemy]{meunier2023emfs}
Etienne Meunier, Anaïs Badoual, and Patrick Bouthemy.
\newblock {EM}-{Driven} {Unsupervised} {Learning} for {Efficient} {Motion}
  {Segmentation}.
\newblock \emph{PAMI}, 2023.

\bibitem[Mildenhall et~al.(2020)Mildenhall, Srinivasan, Tancik, Barron,
  Ramamoorthi, and Ng]{nerf}
Ben Mildenhall, Pratul~P. Srinivasan, Matthew Tancik, Jonathan~T. Barron, Ravi
  Ramamoorthi, and Ren Ng.
\newblock Nerf: Representing scenes as neural radiance fields for view
  synthesis.
\newblock In \emph{ECCV}, 2020.

\bibitem[Ochs and Brox(2011)]{ochs2011object}
Peter Ochs and Thomas Brox.
\newblock Object segmentation in video: a hierarchical variational approach for
  turning point trajectories into dense regions.
\newblock In \emph{ICCV}, 2011.

\bibitem[Ochs et~al.(2014)Ochs, Malik, and Brox]{Ochs2014fbms59}
Peter Ochs, Jitendra Malik, and Thomas Brox.
\newblock Segmentation of {Moving} {Objects} by {Long} {Term} {Video}
  {Analysis}.
\newblock \emph{PAMI}, 2014.

\bibitem[Odobez and Bouthemy(1995)]{odobez1995mrf}
J-M Odobez and Patrick Bouthemy.
\newblock Mrf-based motion segmentation exploiting a 2d motion model robust
  estimation.
\newblock In \emph{ICIP}. IEEE, 1995.

\bibitem[Perazzi et~al.(2016)Perazzi, Pont-Tuset, McWilliams, Van~Gool, Gross,
  and Sorkine-Hornung]{Perazzi2016davis}
Federico Perazzi, Jordi Pont-Tuset, Brian McWilliams, Luc Van~Gool, Markus
  Gross, and Alexander Sorkine-Hornung.
\newblock A {Benchmark} {Dataset} and {Evaluation} {Methodology} for {Video}
  {Object} {Segmentation}.
\newblock In \emph{CVPR}, 2016.

\bibitem[Ranftl et~al.(2021)Ranftl, Bochkovskiy, and Koltun]{dpt2021}
Ren\'{e} Ranftl, Alexey Bochkovskiy, and Vladlen Koltun.
\newblock Vision transformers for dense prediction.
\newblock \emph{ICCV}, 2021.

\bibitem[Rashed et~al.(2019)Rashed, El~Sallab, Yogamani, and
  ElHelw]{Rashed2019mda}
Hazem Rashed, Ahmad El~Sallab, Senthil Yogamani, and Mohamed ElHelw.
\newblock Motion and {Depth} {Augmented} {Semantic} {Segmentation} for
  {Autonomous} {Navigation}.
\newblock In \emph{CVPR Workshop on Visual Odometry and Computer Vision}, 2019.

\bibitem[Ravi et~al.(2024)Ravi, Gabeur, Hu, Hu, Ryali, Ma, Khedr, R{\"a}dle,
  Rolland, Gustafson, Mintun, Pan, Alwala, Carion, Wu, Girshick, Doll{\'a}r,
  and Feichtenhofer]{Ravi2024samv2}
Nikhila Ravi, Valentin Gabeur, Yuan-Ting Hu, Ronghang Hu, Chaitanya Ryali,
  Tengyu Ma, Haitham Khedr, Roman R{\"a}dle, Chloe Rolland, Laura Gustafson,
  Eric Mintun, Junting Pan, Kalyan~Vasudev Alwala, Nicolas Carion, Chao-Yuan
  Wu, Ross Girshick, Piotr Doll{\'a}r, and Christoph Feichtenhofer.
\newblock Sam 2: Segment anything in images and videos.
\newblock \emph{arXiv preprint arXiv:2408.00714}, 2024.

\bibitem[Ren et~al.(2024)Ren, Zhu, Sun, Chen, Pollefeys, and Peng]{onthego}
Weining Ren, Zihan Zhu, Boyang Sun, Jiaqi Chen, Marc Pollefeys, and Songyou
  Peng.
\newblock Nerf on-the-go: Exploiting uncertainty for distractor-free nerfs in
  the wild.
\newblock In \emph{IEEE/CVF Conference on Computer Vision and Pattern
  Recognition (CVPR)}, 2024.

\bibitem[Rombach et~al.(2021)Rombach, Blattmann, Lorenz, Esser, and
  Ommer]{Rombach2021high}
Robin Rombach, Andreas Blattmann, Dominik Lorenz, Patrick Esser, and Bj{\"o}rn
  Ommer.
\newblock High-resolution image synthesis with latent diffusion models.
\newblock In \emph{CVPR}, 2021.

\bibitem[Sabour et~al.(2023)Sabour, Vora, Duckworth, Krasin, Fleet, and
  Tagliasacchi]{sabour2023robustnerf}
Sara Sabour, Suhani Vora, Daniel Duckworth, Ivan Krasin, David~J Fleet, and
  Andrea Tagliasacchi.
\newblock Robustnerf: Ignoring distractors with robust losses.
\newblock In \emph{CVPR}, 2023.

\bibitem[Sabour et~al.(2024)Sabour, Goli, Kopanas, Matthews, Lagun, Guibas,
  Jacobson, Fleet, and Tagliasacchi]{Sabour2024sls}
Sara Sabour, Lily Goli, George Kopanas, Mark Matthews, Dmitry Lagun, Leonidas
  Guibas, Alec Jacobson, David~J. Fleet, and Andrea Tagliasacchi.
\newblock Spotlesssplats: Ignoring distractors in 3d gaussian splatting.
\newblock \emph{arXiv preprint arXiv:2406.20055}, 2024.

\bibitem[Sampson(1982)]{sampson82}
Paul~D. Sampson.
\newblock Fitting conic sections to “very scattered” data: An iterative
  refinement of the bookstein algorithm.
\newblock In \emph{Computer graphics and image processing}, 1982.

\bibitem[Saxena et~al.(2023)Saxena, Herrmann, Hur, Kar, Norouzi, Sun, and
  Fleet]{saxena2023ddvm}
Saurabh Saxena, Charles Herrmann, Junhwa Hur, Abhishek Kar, Mohammad Norouzi,
  Deqing Sun, and David~J. Fleet.
\newblock The surprising effectiveness of diffusion models for optical flow and
  monocular depth estimation.
\newblock In \emph{NeurIPS}, 2023.

\bibitem[Sch\"{o}nberger and Frahm(2016)]{schoenberger2016sfm}
Johannes~Lutz Sch\"{o}nberger and Jan-Michael Frahm.
\newblock Structure-from-motion revisited.
\newblock In \emph{CVPR}, 2016.

\bibitem[Sch\"{o}nberger et~al.(2016)Sch\"{o}nberger, Zheng, Pollefeys, and
  Frahm]{schoenberger2016mvs}
Johannes~Lutz Sch\"{o}nberger, Enliang Zheng, Marc Pollefeys, and Jan-Michael
  Frahm.
\newblock Pixelwise view selection for unstructured multi-view stereo.
\newblock In \emph{ECCV}, 2016.

\bibitem[Sweeney(2015)]{Sweeney2015theiasfm}
Chris Sweeney.
\newblock Theia multiview geometry library: Tutorial \& reference.
\newblock \url{http://theia-sfm.org}, 2015.

\bibitem[Szeliski(2011)]{Szeliski2011cv}
Richard Szeliski.
\newblock \emph{Computer {Vision}: {Algorithms} and {Applications}}.
\newblock Springer, 2011.

\bibitem[Teed and Deng(2020)]{raft2020}
Zachary Teed and Jia Deng.
\newblock {RAFT}: Recurrent all-pairs field transforms for optical flow.
\newblock In \emph{ECCV}, 2020.

\bibitem[Teed and Deng(2021)]{Teed2021droidslam}
Zachary Teed and Jia Deng.
\newblock {DROID}-{SLAM}: {Deep} {Visual} {SLAM} for {Monocular}, {Stereo}, and
  {RGB}-{D} {Cameras}.
\newblock In \emph{NeurIPS}, 2021.

\bibitem[Torr et~al.(1998)Torr, Fitzgibbon, and Zisserman]{torr98}
Phil Torr, Andrew~W. Fitzgibbon, and Andrew Zisserman.
\newblock Maintaining multiple motion model hypotheses over many views to
  recover matching and structure.
\newblock In \emph{ICCV}, 1998.

\bibitem[Wang et~al.(2021)Wang, Luo, Zhang, Zhao, Yin, Wang, Su, Wang, and
  Li]{Wang2021dymslam}
Chenjie Wang, Bin Luo, Yun Zhang, Qing Zhao, Lu Yin, Wei Wang, Xin Su, Yajun
  Wang, and Chengyuan Li.
\newblock {DymSLAM}: {4D} {Dynamic} {Scene} {Reconstruction} {Based} on
  {Geometrical} {Motion} {Segmentation}.
\newblock \emph{IEEE Robotics and Automation Letters}, 2021.

\bibitem[Wang and Adelson(1994)]{wang1994representing}
John~YA Wang and Edward~H Adelson.
\newblock Representing moving images with layers.
\newblock \emph{IEEE transactions on image processing}, 1994.

\bibitem[Wang et~al.(2024{\natexlab{a}})Wang, Leroy, Cabon, Chidlovskii, and
  Revaud]{Wang2023dust3r}
Shuzhe Wang, Vincent Leroy, Yohann Cabon, Boris Chidlovskii, and Jerome Revaud.
\newblock {DUSt3R}: {Geometric} {3D} {Vision} {Made} {Easy}.
\newblock In \emph{CVPR}, 2024{\natexlab{a}}.

\bibitem[Wang et~al.(2024{\natexlab{b}})Wang, Misra, Zeng, Girdhar, and
  Darrell]{Wang2024videocutler}
Xudong Wang, Ishan Misra, Ziyun Zeng, Rohit Girdhar, and Trevor Darrell.
\newblock {VideoCutLER}: {Surprisingly} {Simple} {Unsupervised} {Video}
  {Instance} {Segmentation}.
\newblock In \emph{CVPR}, 2024{\natexlab{b}}.

\bibitem[Weinland et~al.(2011)Weinland, Ronfard, and Boyer]{Weinland2011svb}
Daniel Weinland, Remi Ronfard, and Edmond Boyer.
\newblock A survey of vision-based methods for action representation,
  segmentation and recognition.
\newblock \emph{Computer Vision and Image Understanding}, 2011.

\bibitem[Wilson and Snavely(2014)]{Wilson2014odsfm}
Kyle Wilson and Noah Snavely.
\newblock Robust {Global} {Translations} with {1DSfM}.
\newblock In \emph{ECCV}, 2014.

\bibitem[Wu and Kittler(1993)]{wu1993gradient}
Siu-Fan Wu and Josef Kittler.
\newblock A gradient-based method for general motion estimation and
  segmentation.
\newblock \emph{Journal of Visual Communication and Image Representation},
  1993.

\bibitem[Xie et~al.(2022)Xie, Xie, and Zisserman]{Xie2022oclr}
Junyu Xie, Weidi Xie, and Andrew Zisserman.
\newblock Segmenting {Moving} {Objects} via an {Object}-{Centric} {Layered}
  {Representation}.
\newblock In \emph{NeurIPS}, 2022.

\bibitem[Yang et~al.(2021{\natexlab{a}})Yang, Lamdouar, Lu, Zisserman, and
  Xie]{Yang2021mg}
Charig Yang, Hala Lamdouar, Erika Lu, Andrew Zisserman, and Weidi Xie.
\newblock Self-{Supervised} {Video} {Object} {Segmentation} by {Motion}
  {Grouping}.
\newblock In \emph{ICCV}, 2021{\natexlab{a}}.

\bibitem[Yang et~al.(2021{\natexlab{b}})Yang, Lai, and Soatto]{Yang2021dystab}
Yanchao Yang, Brian Lai, and Stefano Soatto.
\newblock {DyStaB}: {Unsupervised} {Object} {Segmentation} via
  {Dynamic}-{Static} {Bootstrapping}.
\newblock In \emph{CVPR}, 2021{\natexlab{b}}.

\bibitem[Zhang et~al.(2024)Zhang, Herrmann, Hur, Jampani, Darrell, Cole, Sun,
  and Yang]{Zhang2024monst3r}
Junyi Zhang, Charles Herrmann, Junhwa Hur, Varun Jampani, Trevor Darrell,
  Forrester Cole, Deqing Sun, and Ming-Hsuan Yang.
\newblock {MonST3R}: {A} {Simple} {Approach} for {Estimating} {Geometry} in the
  {Presence} of {Motion}.
\newblock \emph{arXiv preprint arXiv:2410.03825}, 2024.

\bibitem[Zhang et~al.(2022)Zhang, Cole, Li, Rubinstein, Snavely, and
  Freeman]{Zhang2022casualsam}
Zhoutong Zhang, Forrester Cole, Zhengqi Li, Michael Rubinstein, Noah Snavely,
  and William~T. Freeman.
\newblock Structure and {Motion} from {Casual} {Videos}.
\newblock In \emph{ECCV}, 2022.

\bibitem[Zhao et~al.(2022)Zhao, Liu, Guo, Wang, and Liu]{Zhao2022particlesfm}
Wang Zhao, Shaohui Liu, Hengkai Guo, Wenping Wang, and Yong-Jin Liu.
\newblock {ParticleSfM}: {Exploiting} {Dense} {Point} {Trajectories}
  for {Localizing} {Moving} {Cameras} in the {Wild}.
\newblock In \emph{ECCV}, 2022.

\end{thebibliography}
}

\clearpage
\appendix
\setcounter{page}{1}
\maketitlesupplementary

\section{Video Examples}
Please refer to our website at \url{https://romosfm.github.io} to view videos of our results. We show video motion segmentation results on \texttt{FBMS59}, \texttt{DAVIS16}, and \texttt{TrackSegv2} compared to \texttt{OCLR-adap}~\cite{Xie2022oclr}. We further show masked video results on Casual Motion dataset and some in-the-wild video samples.

\begin{figure}
\centering
\setlength\tabcolsep{7pt}
\includegraphics[width=0.97\linewidth]{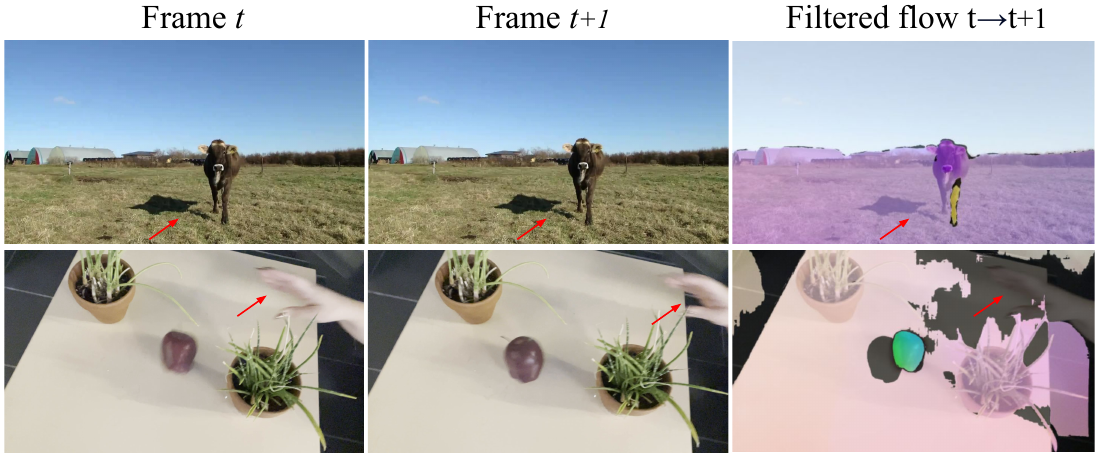} \\

\captionof{figure}{
{\textbf{Optical flow ambiguities in the presence of shadows and occlusion.}} \textbf{Top:} Optical flow of the cow's shadow follows the ground beneath it, although it has a similar movement to the cow. \textbf{Bottom:} The fleetingly appearing arm does not pass optical flow cycle consistency and is completely filtered akin to occluded areas.
}
\label{fig:optical_flow_fails}
\vspace{-0.5cm}
\end{figure}

\section{Optical flow limitations -- Figure \ref{fig:optical_flow_fails}}
\label{sec:optical_flow_limitations}
Despite recent advancements that have made optical flow prediction networks a powerful and versatile tool, there are inherent limitations to optical flow.
One is the ambiguity of flow predictions for shadows~\cite{saxena2023ddvm}.
This can lead to an inability to detect moving shadows as distinct moving entities in our segmentation masks (top of \cref{fig:optical_flow_fails}). 

Another key limitation are objects that appear and disappear almost instantly, such as the arm in our `Table Objects' scene.
These abrupt changes behave similar to occluded areas where the flow is ambiguous and fail the cycle consistency check, rendering nearly all pixels from such objects unusable for our weak inlier/outlier annotations (bottom of  \cref{fig:optical_flow_fails}).

\begin{figure}[t]
\centering
\includegraphics[width=\linewidth]{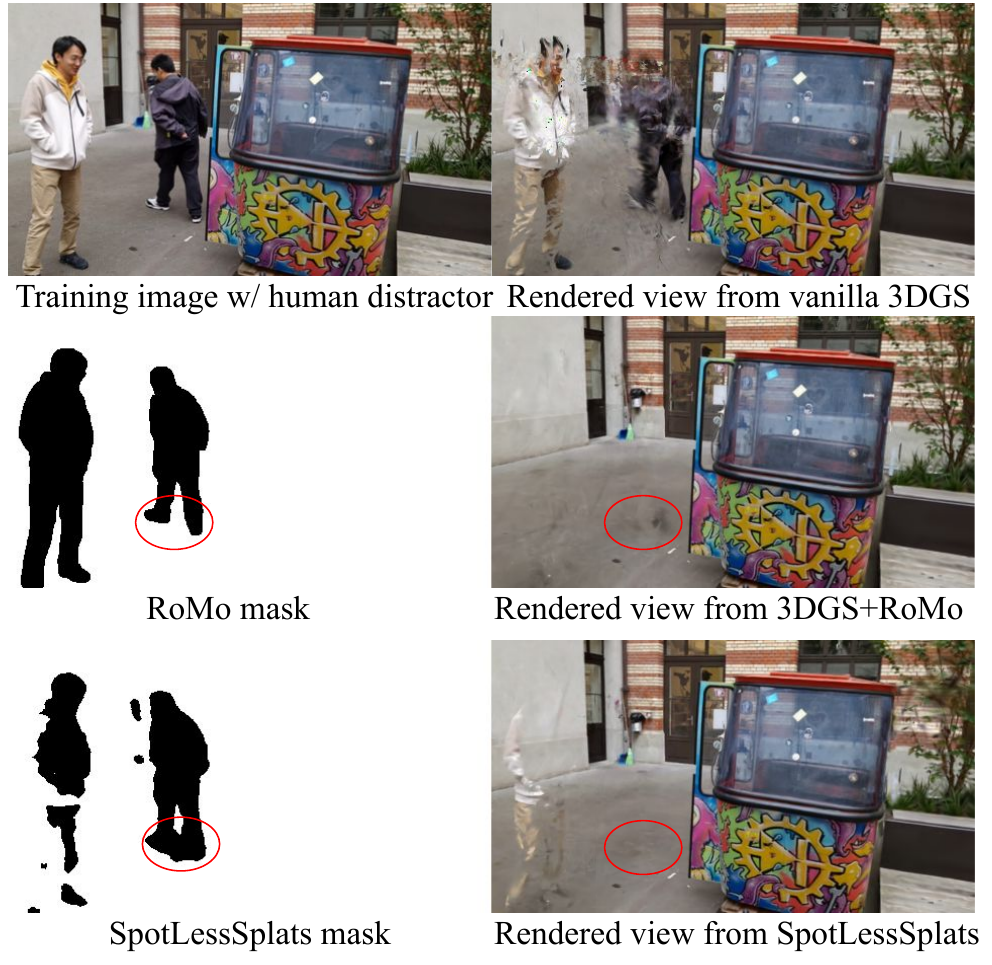}

\caption{  
\textbf{Application of RoMo in 3D optimization} -- with in-the-wild videos, shows that RoMo can completely mask distractor humans in the scenes but fails to capture shadows due to optical flow limitations as described in \cref{sec:optical_flow_limitations}.
}
\label{fig:sls}
\end{figure}

\section{Application in 3D scene optimization with distractors -- Figure \ref{fig:sls}}
Videos, as a collection of images of a scene, can be used to reconstruct the 3D scene using methods like Neural Radiance Fields (NeRFs)~\cite{nerf} or 3D Gaussian Splatting (3DGS)~\cite{3dgs}. 
However, transient inconsistencies, such as passing pedestrians, often violate the static scene assumption of these techniques, appearing as noise in the reconstruction. 
These inconsistencies, referred to as \textit{distractors}, can be filtered out through robust 3D optimization methods, such as those proposed in~\cite{sabour2023robustnerf, onthego, Sabour2024sls}.

\methodshortname  can similarly be be applied to the problem of 3D optimization from such videos by incorporating its motion masks into a standard 3DGS model.
We filter out dynamic pixels from the photometric loss, following the approach in~\citet{sabour2023robustnerf}.
Similar to~\citet{Sabour2024sls}, the structural similarity loss is not utilized in training the 3DGS model.
Qualitative results for this application are presented in~\cref{fig:sls} for the ``patio'' scene from the NeRF On-the-go dataset~\cite{onthego}, which has the temporal order of frames preserved, allowing us to compute optical flow.

Observe that \methodshortname effectively masks moving human distractors in this scene.
We compare against results from SpotLessSplats (SLS)~\cite{Sabour2024sls}, a robust 3D optimization method for 3DGS.
The results show that SLS masks more effectively capture shadows and secondary effects, which \methodshortname misses due to optical flow limitations as discussed earlier. 
However, the results for SLS show leaked distractors in areas of the scene which are sparsely sampled in the training set. This is due to the imbalance of learning rates between the mask predictor of SLS and its 3D model, i.e. the 3D model overfits to the distractor faster than the mask predictor learns its mask.
Adjusting the training schedule to better balance the learning of the mask predictor and the 3DGS model can help mitigate this issue. 
This highlights the inherent challenge of finding an optimal learning rate balance between the two modules in SLS.
In contrast, our approach avoids this problem entirely, as \methodshortname masks are computed as \textit{preprocessing} on the video and provided as input to the 3DGS model optimization.
Because \methodshortname masks operate independently of the 3D optimization pipeline, they can more seamlessly integrate with various 3D reconstruction methods, such as NeRF and 3DGS.
We believe that while our motion masks might not fully capture all inconsistencies for robust 3D optimization, they can serve as a strong initialization for robust masks, which can then be further refined using methods such as SLS.
Furthermore, since \methodshortname does not require camera poses, as many robust 3D optimizations~\cite{sabour2023robustnerf, onthego, Sabour2024sls} do, it can help in cases were SfM pipelines like COLMAP~\cite{schoenberger2016mvs, schoenberger2016sfm} fail due to high distractor rates.

\begin{figure*}[t]
\centering
\setlength\tabcolsep{5pt}
\includegraphics[width=.97\linewidth]{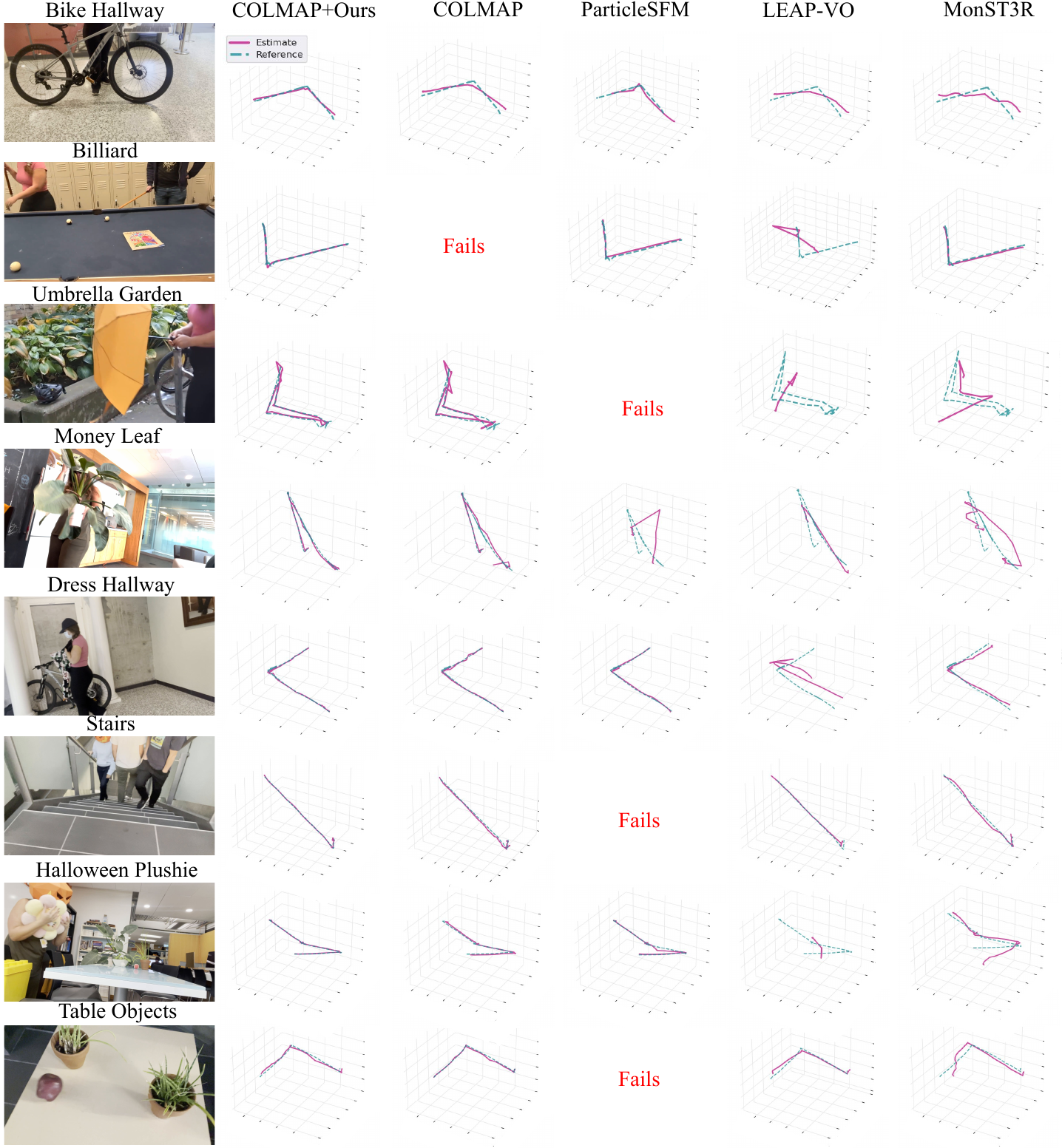}\\
\vspace{0.1cm}
\setlength{\tabcolsep}{5pt}
\resizebox{\linewidth}{!}{
\begin{tabular}{l|ccccc|ccccc|ccccc}
\multirow{2}{*}{} & \multicolumn{5}{c|}{ATE}                                                                                                                                    & \multicolumn{5}{c|}{RPE-T}                                                                                                                                  & \multicolumn{5}{c}{RPE-R}                                                                                                                                  \\ \cline{2-16} 
 $\quad$ Scene & \multicolumn{1}{l}{COLMAP+Ours} & \multicolumn{1}{l}{COLMAP} & \multicolumn{1}{l}{MonST3R} & \multicolumn{1}{l}{LEAP-VO} & \multicolumn{1}{l|}{ParticleSFM} & \multicolumn{1}{l}{COLMAP+Ours} & \multicolumn{1}{l}{COLMAP} & \multicolumn{1}{l}{MonST3R} & \multicolumn{1}{l}{LEAP-VO} & \multicolumn{1}{l|}{ParticleSFM} & \multicolumn{1}{l}{COLMAP+Ours} & \multicolumn{1}{l}{COLMAP} & \multicolumn{1}{l}{MonST3R} & \multicolumn{1}{l}{LEAP-VO} & \multicolumn{1}{l}{ParticleSFM} \\ \hline
Bike Hallway      & \textbf{0.45}                   & 2.17                       & 0.99                        & 0.93                        & 1.68                             & \textbf{0.59}                   & 0.96                       & 0.36                        & 0.64                        & 0.67                             & 1.53                            & 4.46                       & \textbf{0.57}               & 0.72                        & 0.84                            \\
Billiard          & 0.42                            & -                          & 0.34                        & 3.08                        & \textbf{0.31}                    & 0.68                            & -                          & 0.86                        & 1.18                        & \textbf{0.44}                    & 1.42                            & -                          & 1.30                         & 8.19                        & \textbf{0.99}                   \\
Umbrella Garden   & \textbf{0.74}                   & 1.02                       & 2.48                        & 3.15                        & -                                & \textbf{0.66}                   & 1.33                       & 1.13                        & 1.15                        & -                                & \textbf{2.30}                   & \textbf{2.30}              & 3.82                        & 6.39                        & -                               \\
Money Leaf        & \textbf{0.59}                   & 0.65                       & 2.23                        & 2.55                        & 1.84                             & \textbf{0.97}                   & 0.98                       & 1.01                        & 1.09                        & 1.36                             & \textbf{4.53}                   & 4.56                       & 4.67                        & 6.65                        & 5.9                             \\
Dress Hallway     & \textbf{0.27}                   & 0.36                       & 0.62                        & 2.42                        & 0.29                             & \textbf{0.37}                   & 0.56                       & 0.58                        & 0.64                        & 0.40                             & \textbf{1.97}                   & 2.56                       & 2.83                        & 1.98                        & 1.79                            \\
Stairs            & \textbf{0.22}                   & 0.31                       & 0.31                        & 0.58                        & -                                & \textbf{0.23}                   & 0.28                       & 0.50                        & 0.45                        & -                                & \textbf{0.12}                   & 0.15                       & 0.74                        & 0.30                        & -                               \\
Halloween Plushie & \textbf{0.21}                   & \textbf{0.21}                       & 1.05                        & 3.5                         & 0.25                             & \textbf{0.18}                   & 0.19                       & 0.76                        & 0.82                        & 0.20                             & \textbf{0.43}                   & 0.45                       & 1.64                        & 5.93                        & 0.46                            \\
Table Objects     & 0.35                            & \textbf{0.16}              & 0.74                        & 0.62                        & -                                & 0.17                            & \textbf{0.11}              & 0.60                        & 0.22                        & -                                & \textbf{0.45}                   & 0.72                       & 2.43                        & 0.8                         & -                              
\end{tabular}
}%
\captionof{figure}{
\textbf{Detailed results on ``Casual Motion''} -- 
show that our method can be paired with a bundle adjustment technique (COLMAP~\cite{schoenberger2016sfm}) to make it more robust to dynamic scenes, often outperforming SoTA methods for camera estimation on such scenes. 
}
\label{fig:ourdatasetdetail}
\end{figure*}

\section{Results on ``Casual Motion'' -- Figure \ref{fig:ourdatasetdetail}}
\label{sec:casual_motion}

Figure~\ref{fig:ourdatasetdetail} presents a more detailed breakdown of results from our ``Casual Motion'' dataset (main paper Figure 9).
It illustrates that supervised baselines, which rely heavily on synthetic data, have less reliable estimates of camera pose compared to classic camera estimation methods like \texttt{COLMAP}~\cite{schoenberger2016mvs, schoenberger2016sfm}.
The `Money Leaf' scene exemplifies significant challenges for \texttt{ParticleSFM}~\cite{Zhao2022particlesfm}, \texttt{LEAP-VO}~\cite{Chen2024leapvo}, and \texttt{MonST3R}~\cite{Zhang2024monst3r}, all of which produce notably inferior results compared to \texttt{COLMAP}.
In contrast, our method leverages \texttt{COLMAP}'s strength as a robust camera pose estimator while addressing its limitations.
This enhancement is evident both quantitatively and qualitatively, particularly at the beginnings and ends of trajectories.
In these regions, where slower camera movements with smaller translation are overshadowed by the larger motions of dynamic objects, \texttt{COLMAP}'s estimates often falter. 
Our approach corrects these errors effectively by incorporating dynamic masks.

\begin{figure*}[t]
\centering
\setlength\tabcolsep{5pt}
\includegraphics[width=.9\linewidth]{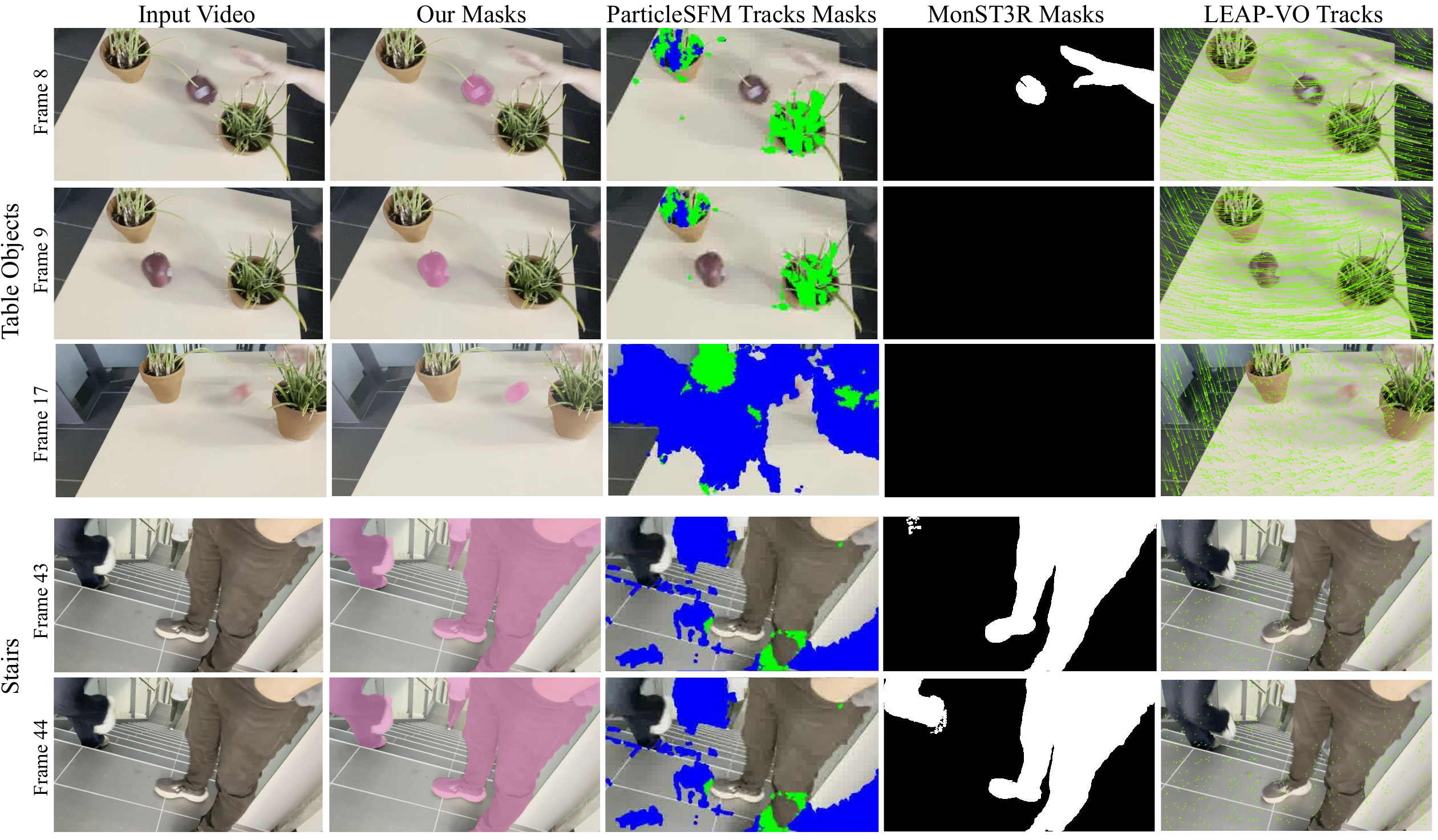}\\

\captionof{figure}{
\textbf{Detailed results on ParticleSFM~\cite{Zhao2022particlesfm} failing scenes} -- 
shows that over masking static regions can lead to bundle adjustment failure. Moreover, sparse tracks on highly occluded frames can lead to failure. 
}
\label{fig:psfm}
\end{figure*}

\section{Failure scenes of ParticleSFM -- Figure \ref{fig:psfm}}

Figure~\ref{fig:psfm} presents a detailed comparison of camera pose estimation baselines on scenes where \texttt{ParticleSFM} struggles. 
The `Table Objects' scene is particularly challenging due to rapid camera and rapid object movements, which result in motion blur and sparse dynamic objects.
These factors make masking difficult for all methods, including ours.
\texttt{COLMAP} is generally robust to this scene because the movements, though rapid, are temporally sparse.
Poor masking however, can lead to failures in the robust baselines.
Qualitative results show that \texttt{ParticleSFM}~\cite{Zhao2022particlesfm} focuses its detected tracks (blue and green) and filtered dynamic tracks (green) on the static flowerpots, which provide texture and reliable cues for bundle adjustment in an otherwise plain-textured scene.
This incorrect masking causes \texttt{ParticleSFM} to completely fail at camera estimation. \texttt{MonST3R}~\cite{Zhang2024monst3r} produces good masks in some frames but fails with empty masks in others.
\texttt{LEAP-VO}~\cite{Chen2024leapvo} shows no evidence of filtering tracks associated with dynamic objects (green arrows).
Our method partially fails to detect the fleetingly appearing arm but successfully masks out the moving fruits even under heavy blur.

The `Stairs' scene presents a highly occluded environment.
\texttt{ParticleSFM} fails to estimate camera poses for the final frames with the most occlusions, likely due to the sparsity of remaining tracks (blue region in Figure~\ref{fig:psfm}).
\texttt{MonST3R} occasionally misses moving people, and \texttt{LEAP-VO} does not filter tracks of dynamic objects.
In contrast, \methodshortname fully masks the dynamic people in this scene.

Finally, in the `Umbrella Garden' scene, \texttt{ParticleSFM} fails to find sufficient tracks due to the high occlusion rate during its initial stage, leading to a complete failure.


\section{Detailed results on MPI Sintel -- Table \ref{fig:sintel_detail}}
\Cref{fig:sintel_detail} presents a per-scene breakdown of results on MPI Sintel for both our method and unmasked TAPIR~\cite{Doersch2023tapir} tracks used with TheiaSfM~\cite{Sweeney2015theiasfm}.

\begin{table*}[t]
\centering
\setlength\tabcolsep{7pt}
\vspace{-.2cm}
\setlength{\tabcolsep}{24pt}
\resizebox{\linewidth}{!}{
\begin{tabular}{c|ccc||ccc}
      \multirow{2}{*}{Scene}      & \multicolumn{3}{c||}{Our Masks + TAPIR tracks + TheiaSFM}                          & \multicolumn{3}{c}{TAPIR tracks + TheiaSFM}                         \\ \cline{2-7}
       & \multicolumn{1}{c|}{ATE}   & \multicolumn{1}{c|}{RPE (T)}        & RPE (R)        & \multicolumn{1}{c|}{ATE}   & \multicolumn{1}{c|}{RPE (T)} & RPE (R) \\ \hline
alley\_2    & \multicolumn{1}{c|}{0.001} & \multicolumn{1}{c|}{0.001}          & 0.018          & \multicolumn{1}{c|}{0.001} & \multicolumn{1}{c|}{0.001}   & 0.020   \\ \hline
ambush\_4   & \multicolumn{1}{c|}{0.014} & \multicolumn{1}{c|}{0.015}          & 0.188          & \multicolumn{1}{c|}{0.017} & \multicolumn{1}{c|}{0.014}   & 0.159   \\ \hline
ambush\_5   & \multicolumn{1}{c|}{0.004} & \multicolumn{1}{c|}{0.004}          & 0.068          & \multicolumn{1}{c|}{0.037} & \multicolumn{1}{c|}{0.027}   & 0.750   \\ \hline
ambush\_6   & \multicolumn{1}{c|}{0.003} & \multicolumn{1}{c|}{0.002}          & 0.047          & \multicolumn{1}{c|}{0.150} & \multicolumn{1}{c|}{0.090}   & 1.802   \\ \hline
cave\_2     & \multicolumn{1}{c|}{0.773} & \multicolumn{1}{c|}{0.176}          & 0.626          & \multicolumn{1}{c|}{0.782} & \multicolumn{1}{c|}{0.170}   & 0.683   \\ \hline
cave\_4     & \multicolumn{1}{c|}{0.005} & \multicolumn{1}{c|}{0.003}          & 0.019          & \multicolumn{1}{c|}{0.078} & \multicolumn{1}{c|}{0.046}   & 0.283   \\ \hline
market\_2   & \multicolumn{1}{c|}{0.014} & \multicolumn{1}{c|}{0.012}          & 0.112          & \multicolumn{1}{c|}{0.068} & \multicolumn{1}{c|}{0.028}   & 8.483   \\ \hline
market\_5   & \multicolumn{1}{c|}{0.010}  & \multicolumn{1}{c|}{0.003}          & 0.027          & \multicolumn{1}{c|}{0.012} & \multicolumn{1}{c|}{0.004}   & 0.029   \\ \hline
market\_6   & \multicolumn{1}{c|}{0.006} & \multicolumn{1}{c|}{0.005}          & 0.037          & \multicolumn{1}{c|}{0.051} & \multicolumn{1}{c|}{0.022}   & 0.800   \\ \hline
shaman\_3   & \multicolumn{1}{c|}{0.001} & \multicolumn{1}{c|}{0.001}          & 0.213          & \multicolumn{1}{c|}{0.005} & \multicolumn{1}{c|}{0.003}   & 0.680   \\ \hline
sleeping\_1 & \multicolumn{1}{c|}{0.009} & \multicolumn{1}{c|}{0.009}          & 0.898          & \multicolumn{1}{c|}{0.011} & \multicolumn{1}{c|}{0.013}   & 1.267   \\ \hline
sleeping\_2 & \multicolumn{1}{c|}{0.001} & \multicolumn{1}{c|}{0.001}          & 0.026          & \multicolumn{1}{c|}{0.001} & \multicolumn{1}{c|}{0.001}   & 0.026   \\ \hline
temple\_2   & \multicolumn{1}{c|}{0.002} & \multicolumn{1}{c|}{0.002}          & 0.009          & \multicolumn{1}{c|}{0.002} & \multicolumn{1}{c|}{0.002}   & 0.008   \\ \hline
temple\_3   & \multicolumn{1}{c|}{0.456} & \multicolumn{1}{c|}{0.128}          & 0.743          & \multicolumn{1}{c|}{0.626} & \multicolumn{1}{c|}{0.204}   & 1.452   \\ \hline \hline
Avg         & \multicolumn{1}{c|}{\textbf{0.093}} & \multicolumn{1}{c|}{\textbf{0.026}} & \textbf{0.217} & \multicolumn{1}{c|}{0.132} & \multicolumn{1}{c|}{0.045}   & 1.175  
\end{tabular}

}%
\captionof{table}{
Per scene breakdown of MPI Sintel~\cite{Butler2012sintel} results.
}
\vspace*{-0.1cm}
\label{fig:sintel_detail}
\end{table*}

\end{document}